\documentclass{article}

\usepackage{soul}

\usepackage{microtype}
\usepackage{graphicx}
\usepackage{subfigure}
\usepackage{booktabs} %
\usepackage{wrapfig}
\usepackage{multirow}

\usepackage[pdfusetitle]{hyperref}

\hypersetup{hidelinks}

\PassOptionsToPackage{square,numbers}{natbib}
\usepackage[final]{Neurips/neurips_2025}
\usepackage{algpseudocode}
\usepackage{algorithm}

\usepackage{amsmath}
\usepackage{amssymb}
\usepackage{mathtools}
\usepackage{amsthm}
\usepackage{thmtools}

\usepackage[capitalize,noabbrev]{cleveref}

\usepackage[utf8]{inputenc} %
\usepackage[T1]{fontenc}    %
\usepackage{hyperref}       %
\usepackage{url}            %
\usepackage{booktabs}       %
\usepackage{amsfonts}       %
\usepackage{nicefrac}       %
\usepackage{microtype}      %
\usepackage{xcolor}         %
\usepackage{xkcdcolors}

\hypersetup{
    colorlinks=true,
    linkcolor=xkcdDarkBlue,      %
    filecolor=xkcdDarkPurple,    %
    urlcolor=xkcdDarkBlue,           %
    citecolor=xkcdDarkBlue,         %
    anchorcolor=black
}

\theoremstyle{plain}
\newtheorem{theorem}{Theorem}[section]

\newtheorem{lemma}[theorem]{Lemma}
\newtheorem{corollary}[theorem]{Corollary}
\theoremstyle{definition}
\newtheorem{definition}[theorem]{Definition}

\theoremstyle{remark}

\usepackage[dvipsnames]{xcolor}

\usepackage{graphicx}
\usepackage{graphics}
\usepackage{booktabs}
\usepackage{pdflscape}
\usepackage{array}
\usepackage{siunitx}
\newcolumntype{H}{>{\setbox0=\hbox\bgroup}c<{\egroup}@{}} %

\usepackage{amsmath,amsfonts,bm}

\newcommand{\prob}[1]{\mathbb P\left[#1 \right]}

\mathchardef\mhyphen="2D

\counterwithin*{theorem}{section}

\theoremstyle{definition}

\theoremstyle{definition}  

\crefname{axiom}{Axiom}{Axioms}

\def\eqref#1{equation~\ref{#1}}

\def\1{\bm{1}}

\DeclareMathOperator{\hash}{HASH}

\def\mA{{\bm{A}}}

\def\mI{{\bm{I}}}

\def\mX{{\bm{X}}}

\DeclareMathAlphabet{\mathsfit}{\encodingdefault}{\sfdefault}{m}{sl}
\SetMathAlphabet{\mathsfit}{bold}{\encodingdefault}{\sfdefault}{bx}{n}

\def\sL{{\mathbb{L}}}

\def\sS{{\mathbb{S}}}

\DeclarePairedDelimiterX{\infdivx}[2]{[}{]}{%
  #1\;\delimsize\|\;#2%
}

\crefname{figure}{Fig.}{Figs.}
\crefname{table}{Tab.}{Tabs.}
\crefname{section}{\S}{\S\S}
\crefname{chapter}{Ch.}{Chs.}
\crefname{appendix}{App.}{Apps.}
\crefname{equation}{Eq.}{Eqs.}
\crefname{theorem}{Thm.}{Thms.}

\newcommand{\hlc}[2]{\sethlcolor{#1}\hl{#2}}
\definecolor{hl1}{RGB}{255, 235, 205}   %
\definecolor{lossy}{RGB}{255, 235, 205} 

\definecolor{hl2}{RGB}{204, 255, 204}   %
\definecolor{binary}{RGB}{204, 255, 204}   %

\definecolor{hl3}{RGB}{204, 229, 255}   %
\definecolor{global}{RGB}{204, 229, 255} 

\definecolor{hl4}{RGB}{255, 204, 229}   %
\definecolor{graphs}{RGB}{255, 204, 229}   %

\definecolor{hl5}{RGB}{230, 220, 255}   %
\definecolor{task}{RGB}{230, 220, 255}   %

\definecolor{hl6}{RGB}{224, 255, 255}   %

\newcommand{\mathhighlight}[2]{\colorbox{#1}{$\displaystyle #2$}}

\def\graphs{{\mathcal{G}}}
\def\graphsV{{\graphs_\mathcal{V}}}
\def\graph{{G}}
\def\vertices{{\mathcal{V}_\graph}}
\def\features{\mX}
\def\adjacency{\mA}

\def\edges{{\mathcal{E}_\graph}}
\def\graphT{{\tilde{\graph}}}

\def\neighbor{{N}}
\def\adjacency{{{\mA}}}

\def\AGG{{\texttt{agg}}}

\def\msg{{\texttt{msg}}}
\def\upd{{\texttt{upd}}}

\def\model{{M}}
\def\modelA{{\mathcal{M}}}
\def\modelAS{{\modelA_S}}

\def\normInf{{\mI}}

\def\WL{{\textup{WL}}}

\def\ourWL{\textup{lossyWL}}
\def\ours{{\textup{MPC}}}

\def\WLComp{\textup{WLC}}

\def\prob{\mathbb{P}}

\newcommand*{\ldbrace}{\{\mskip-5mu\{}
\newcommand*{\rdbrace}{\}\mskip-5mu\}}

\title{What Expressivity Theory Misses:\\ Message Passing Complexity for GNNs}

\author{%
  Niklas Kemper
  \quad \quad
  Tom Wollschläger
  \quad \quad
  Stephan Günnemann \\
  School of Computation, Information and Technology \& Munich Data Science Institute \\
  Technical University of Munich \\
  \texttt{\{niklas.kemper, t.wollschläger, s.guennemann\}@tum.de} 
}

\begin{document}

\maketitle

\begin{abstract}
  Expressivity theory, characterizing which graphs a GNN can distinguish, has become the predominant framework for analyzing GNNs, with new models striving for higher expressivity. However, we argue that this focus is misguided: First, higher expressivity is not necessary for most real-world tasks as these tasks rarely require expressivity beyond the basic WL test. Second, expressivity theory's binary characterization and idealized assumptions fail to reflect GNNs' \emph{practical} capabilities.
To overcome these limitations, we propose Message Passing Complexity (\ours{}): a continuous measure that quantifies the difficulty for a GNN architecture to solve a given task through message passing.
\ours{} captures practical limitations like over-squashing while preserving theoretical impossibility results from expressivity theory, effectively narrowing the gap between theory and practice.
Through extensive validation on fundamental GNN tasks, we show that \ours{}'s theoretical predictions correlate with empirical performance, successfully explaining architectural successes and failures. Thereby, \ours{} advances beyond expressivity theory to provide a more powerful and nuanced framework for understanding and improving GNN architectures.

\end{abstract}

\section{Introduction}
\label{sec:introduction}
From weather forecasting to drug design, Graph Neural Networks (GNNs) have shown remarkable success across diverse applications. 
However, the seminal works by \citet{morris} and \citet{xu} revealed a key limitation of standard Message Passing Neural Networks (MPNNs): 
Their ability to distinguish non-isomorphic graphs can be bounded by the Weisfeiler-Leman (WL) graph isomorphism test \cite{wl}.

In response, significant research effort has focused on developing more expressive architectures that surpass the WL test \cite{jegelka_survey}.
The underlying hypothesis is that increased expressivity translates to better empirical performance, with improved benchmark results often attributed to higher expressivity \cite{fragNet}. Isomorphism-based (Iso) expressivity theory, which characterizes the sets of graphs an architecture can distinguish through the WL test and its extensions, has thus become the predominant framework for analyzing MPNNs.

We argue that this focus on iso expressivity is misguided. 
While it provides valuable impossibility results, we identify two key limitations that prevent it from explaining real-world MPNN performance. First, higher expressivity is often not necessary for real-world tasks: almost all graphs in standard benchmarks are already distinguishable by the basic WL test, making it unclear why higher expressivity would improve performance \cite{wl_enough}. Second, iso expressivity theory fails to capture practical model capabilities. It assumes unrealistic conditions like lossless information propagation over unbounded layers, ignoring practical limitations like over-squashing \cite{oversquashing_1}. Moreover, its binary view (can vs. cannot distinguish) offers no insight into the relative difficulty of learning specific real-world tasks.

To address these limitations, we propose shifting from binary expressivity tests to a continuous complexity measure, MPC, that quantifies the message-passing complexity of \emph{arbitrary} tasks for a given architecture. MPC builds upon a novel probabilistic WL test. It captures practical limitations of MPNNs, such as under-reaching \cite{under_reaching} and over-squashing \cite{oversquashing_1}, which are known to hinder empirical performance, while preserving impossibility results from iso expressivity theory—effectively narrowing the gap between theory and practice.

Through extensive validation, we show that trends in MPC complexity align with empirical performance across a range of fundamental graph tasks.\footnote{Find our implementation at \url{https://www.cs.cit.tum.de/daml/message-passing-complexity/}} Notably, success is determined not by iso expressivity but by architectural choices that minimize complexity for specific tasks. For instance, a simple GCN with virtual node outperforms strictly more expressive, higher-order models at long-range tasks, as MPC correctly predicts. By providing a quantitative measure of architectural capabilities for specific tasks, MPC both reveals current model limitations and offers clear optimization targets for future architectural innovations, shifting focus from maximizing expressivity to minimizing task-specific complexity.

In summary, our key contributions are:  
\begin{itemize}  
    \item We identify limitations of iso expressivity theory that prevent it from explaining MPNN performance in real-world tasks (\cref{sec:expressivity}). 
    \item We introduce \ours{}, a continuous message-passing complexity measure rooted in a novel probabilistic WL test that characterizes task-specific difficulty.  \ours{} captures existing MPNN limitations, such as over-smoothing and under-reaching, while retaining impossibility results from iso expressivity theory (\cref{sec:complexity}).  
    \item We extensively validate \ours{}, showing its consistency with empirical performance and its superiority over classical expressivity theory in explaining real-world MPNN behavior (\cref{sec:evaluation}).
\end{itemize}

\section{Background}
\label{sec:background}
\textbf{Notation}\quad
Let $\graph = (\vertices, \edges, \features)$ denote an (undirected) graph with nodes $\vertices$, edges $\edges$, features $\features_v$ for $v \in \vertices$ and adjacency matrix $\adjacency$. With ${\tilde{\adjacency} = \adjacency + \mathbf{I}}$, define the influence matrix ${\normInf_{u v} := \tilde{\adjacency}_{u v} / \sum_{w} \tilde{\adjacency}_{u w}}$ as the normalized adjacency (with self-loops). Let $d_\graph(u,v)$ be the shortest path distance between $u$ and $v$ and $\neighbor_\graph(v)$ be the set of neighbors of $v$.
Let $\graphs$ denote a set of graphs and $\graphs^{*}$ the set of all graphs. Let $\graphsV= \{(\graph,u) \mid \graph \in \graphs, u \in \vertices \} $ denote the set of graph-node pairs. We will mainly consider node-level functions or tasks of the form $f: \graphsV \to \mathbb{R}^k$. For brevity, we will often write $f_v(\graph)$ for $f(\graph, v)$. We define $\log(0) = -\infty$ and $\ldbrace . \rdbrace$ is a multiset.
Let $\WL^l: \graphsV \to \mathbb{R}^{k}$ denote the color assignment of the $l$-th round of the WL test, defined as:
$\WL^0_v = \features_v$ and $\WL^l_v = \hash\left(\WL^{l-1}_{v}, \ldbrace \WL^{l-1}_u \mid u \in \neighbor_\graph(v) \rdbrace \right)$.

\textbf{Message Passing Framework}\quad
Standard MPNNs $\modelAS$ have hidden representations $h_v^l$ for each node $v$ which are updated iteratively at each layer $l \in \{1, \dots, L\}$ by aggregating messages $m_{w \to v}^l$ from neighboring nodes $w$ in the graph $\graph$:
\begin{align*}
      m_{w \to v}^l := \begin{cases} \msg_0^l(h_w^{l-1}) & \text{if}\:\: w=v \\ \msg_1^l(h_w^{l-1}) & \text{else} \end{cases} \quad \text{and}\quad h_v^l := \upd^l (\AGG \left ( \ldbrace  m_{w \to v}^l \mid w \in \neighbor_\graph(v) \cup \{v\} \rdbrace \right ) )
\label{eq:mpnn2}
 \end{align*}
Here,  $\smash{\msg^l_0}$, $\smash{\msg^l_1}$, and $\smash{\upd^l}$ can be arbitrary (learned) functions, often MLPs. Typical choices for the aggregation function $\AGG{}$ are mean or sum. 
We differentiate between a \emph{model architecture} $\modelA$ that only specifies which nodes exchange messages (abstracting away from the choice of $\msg$, $\upd$, and $\AGG$), and a (learned) \emph{model instantiation} $\model \in \modelA$ with fully specified $\msg$, $\upd$ and $\AGG$ functions. Standard MPNNs $\modelAS$ perform message passing directly on the input graph $\graph$. In contrast, more recent architectures propagate messages on a transformed message passing (MP) graph $\smash{\graphT} = t(\graph)$ \cite{mp_all_the_way}. They introduce modifications like additional virtual nodes \cite{vn}, rewired edges \cite{oversquashing_curvature}, or higher-order graphs \cite{fragNet}. For simplicity, we mainly focus on standard MPNNs $\modelAS$ in the main text, and defer a general framework encompassing architectures $\modelA$ with arbitrary MP graphs $\smash{\graphT}$ to \cref{app:mpnn}.

\section{Limitations of Expressivity Theory}
\label{sec:expressivity}
\begin{wrapfigure}{r}{0.395\textwidth} %
\centering
\vspace{-0.3cm}
\includegraphics[width=0.95\linewidth]{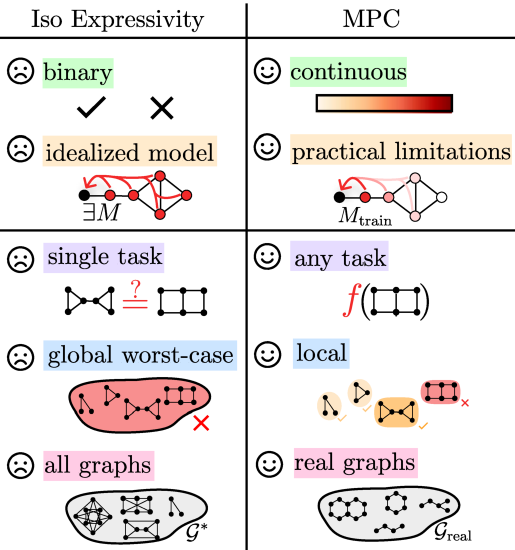} %
\vspace{-0cm}
\caption{Limitations of iso expressivity vs. benefits of \ours{}. Top: Iso expressivity gives an idealized, binary view that misses GNNs’ practical capabilities. Bottom: Limited expressivity rarely restricts real-world performance as it focuses on worst-case graphs and tasks.}
\label{fig:overview}
\vspace{-0.1cm}
\end{wrapfigure}

MPNN architectures differ in their theoretical capacity to solve graph tasks, such as detecting specific substructures \cite{substructure_wl}. Most prior expressivity theory captures differences by considering an architecture's ability to distinguish non-isomorphic graphs relative to a reference isomorphism test $\alpha$ \cite{cinpp,cw,fragNet,overview_wl}. Formally:
\begin{definition}[Iso Expressivity]
\label{def:iso_expressivity}
Let $\alpha$ be a \hlc{hl5}{graph isomorphism test}. An architecture $\modelA$ is at least as expressive as $\alpha$ if \hlc{hl1}{$\exists \model \in {\modelA}$} such that \hlc{hl3}{$\forall G, G'$} $\in$ \hlc{hl4}{$\graphs^*$}:
\begin{equation}
\mathhighlight{hl2}{\model(\graph) = \model(\graph')} \implies \alpha(G) = \alpha(G').
\end{equation}
 Contrarily, $\modelA$ is at most as expressive as $\alpha$ if $\forall M \in \modelA$ and \hlc{hl3}{$\forall G, G'$} $\in$ \hlc{hl4}{$\graphs^*$}
\begin{equation}
\alpha(G) = \alpha(G')\implies \model(\graph) = \model(\graph'). 
\end{equation}
\end{definition}
Standard MPNNs $\modelAS$ are at most as expressive as the WL test \cite{morris,xu}.
In response, recent GNN research has focused on developing architectures that surpass the WL test in expressivity. Their strong performance on benchmarks such as ZINC \cite{cw,gsn,fwl,fragNet} is often motivated by and attributed to their higher iso expressivity \cite{fragNet,gsn,fwl}. This builds on the premise that theoretical iso expressivity correlates with empirical performance. Consequently, this line of expressivity research rests on two fundamental assumptions, both of which we show to be problematic (\cref{fig:overview}):
\begin{enumerate}
    \item \textit{Iso expressivity theory accurately describes the practical capabilities of trained MPNNs.} The theoretical higher expressivity of higher-order MPNNs translates to a higher practical capability to learn more complex tasks, contributing to improved empirical performance.
    \label{item:assumption1}
    \item \textit{Iso expressivity beyond WL is important for real-world performance.} The limited expressivity of standard MPNNs restricts their performance on real-world tasks.
    \label{item:assumption2} 
\end{enumerate}

We first critique Assumption \ref{item:assumption1}, showing that iso expressivity theory relies on idealized assumptions and provides incomplete insights about practical MPNN capabilities. Positive expressivity results only guarantee the existence of a model instance within the architecture that is more expressive than $\alpha$. These proofs typically rely on a maximally expressive model $M^*$ with injective \upd{} functions and an unbounded number of layers \cite{cw}: conditions under which every node can access the complete information from all other nodes.
This idealized setup \hlc{hl1}{ignores fundamental limitations of real-world MPNNs}. In practice, \hlc{hl1}{message passing is lossy}: information is often bottlenecked (over-squashing \cite{oversquashing_1}), blurred (over-smoothing \cite{oversmoothing}), or simply out of reach due to shallow depth (under-reaching \cite{under_reaching}). As a result, real-world MPNNs often fail to propagate information even between nearby nodes, let alone replicate the idealized behavior assumed in expressivity theory.
Second, expressivity theory provides strictly \hlc{hl2}{binary results} \cite{future_directions}: can vs. cannot distinguish, with \hlc{hl2}{no indication of difficulty}. This binary lens cannot account for the wide variation in empirical performance between architectures of equal expressivity on the same task. For example, adding a virtual node does not increase iso expressivity \cite{fragNet} but often leads to performance gains on long-range tasks \cite{vn_empirical}.
Additionally, \emph{between} theoretically solvable tasks, iso expressivity theory provides no insight into their relative difficulty. Some tasks might be trivially learnable in practice while others may be practically impossible to learn with finite data and training time.
In summary, the assumptions behind positive expressivity results diverge sharply from the realities of real-world MPNNs, and the binary view can offer only limited insights for practice. Our empirical results in \cref{sec:evaluation} reinforce this, showing that even highly expressive MPNN architectures struggle with elementary tasks like maintaining initial node features. This highlights a gap between positive iso expressivity results and practical performance.

We now challenge Assumption~\ref{item:assumption2}, asking whether negative expressivity results reflect meaningful limitations in practice. Iso expressivity theory considers \hlc{hl5}{only the hardest possible task: distinguishing all non-isomorphic graphs}. However, many real-world applications, such as social network analysis or relational learning, often require only aggregating (local) information. In these cases, the inability to distinguish certain non-isomorphic graphs does not constitute a meaningful limitation in practice. By focusing exclusively on this worst-case task, expressivity theory overlooks the specific requirements of practical problems.
 Second, expressivity theory makes \hlc{hl3}{global statements over the set of graphs} $\graphs^*$,  demonstrating a lack of expressivity through single, carefully constructed counterexamples \cite{fragNet,cinpp,cw}. However, these counterexamples are rare: the probability of encountering WL equivalent graphs in random graphs approaches 0 \cite{random_wl}, and real-world graphs often carry rich node features that further break WL equivalences. Hence,  the mere existence of theoretical counterexamples provides limited insight into a model's practical performance on large, diverse datasets.
  Lastly, iso expressivity theory is \hlc{hl4}{graph-family agnostic}. Its results are derived for the set of \hlc{hl4}{all possible graphs $\smash{\graphs^*}$}, whereas real-world applications typically involve restricted graph families. For example, molecular datasets primarily contain planar graphs with bounded degrees. Tasks infeasible over all graphs may become solvable for standard MPNNs within these restricted graph families \cite{planar,gnns_can_count}. Moreover, as shown in \cref{tab:wl}, the basic \emph{WL test already distinguishes almost all graph pairs in popular benchmarks} across different domains \cite{wl_enough}. Therefore, iso expressivity theory cannot explain why expressivity beyond the WL test would benefit these real-world tasks or account for the performance differences between architectures.

Our analysis shows that iso expressivity theory, while valuable for establishing theoretical limitations, provides limited insights for practical MPNN applications. It assumes idealized conditions and focuses on theoretical worst cases, leading to a disconnect with practice: high iso expressivity does not imply good performance (contradicting Assumption \ref{item:assumption1}), and limited iso expressivity does not imply poor performance (contradicting Assumption \ref{item:assumption2}). While alternative expressivity approaches such as logic-based characterizations address certain limitations of iso expressivity (see \cref{sec:related_work}), none can characterize the varying degrees of practical learning difficulty encountered in real-world tasks. These findings suggest that pursuing higher expressivity alone may be misguided, highlighting the need for a framework that captures theoretical limitations \emph{and} practical learning challenges.

\section{Message Passing Complexity}
\label{sec:complexity}
To move beyond the limitations of expressivity theory, we propose our continuous, task-specific message passing complexity (\ours{}). Unlike iso expressivity theory, which only asks whether an MPNN can distinguish certain graphs in theory, \ours{} quantifies \emph{how difficult} it is for messages to propagate through a graph \emph{to solve a given task}. By accounting for the inherently lossy information propagation of real-world MPNNs, \ours{} can explain practical performance trends that expressivity theory misses. 
While many complexity measures exist in machine learning and theoretical computer science, they fail to address this unique challenge of MPNNs: propagating information effectively across graph structures. \ours{} specifically isolates this challenge from other general sources of difficulty in machine learning—for instance, learning high-degree polynomials is a well-understood difficulty not specific to graph learning. For clarity, we focus on node-level tasks $f_v$ and standard MPNNs $\modelAS$ in this section, with generalizations to broader architectures and tasks provided in \cref{app:mpnn}.

As an initial step, we define a \hlc{global}{local} \hlc{task}{task-specific} complexity measure based on the WL test that considers \hlc{global}{individual graphs} and \hlc{task}{individual tasks} rather than making global worst-case statements over all graphs and tasks. For this, we first need to formalize when one individual function output provides sufficient information to determine another (\cref{fig:finegrained_example}):
 \begin{definition}
 \label{def:fine_grained}
    Let $\alpha: X \to Y$, $\beta: X \to Z$ be two functions. Let $x \in X$ be fixed. Then, $\beta(x)$ \emph{can be deduced} from $\alpha(x)$, $\alpha(x) \vDash_X \beta(x)$, iff 
    \[\forall x' \in X: \:\alpha(x) = \alpha(x') \Rightarrow \beta(x) = \beta(x').\]
    If $\alpha(x) \vDash_X \beta(x)$ for all $x \in X$, we write $\alpha \vDash_X \beta$ and say $\alpha$ is more \emph{fine-grained} than $\beta$.
\end{definition}
Intuitively, $\alpha(x) \vDash_X \beta(x)$ means that $\alpha(x)$ provides sufficient information to uniquely determine $\beta(x)$.
Using this, we can define a first complexity measure dependent on a specific task $f_v$ and individual graph $\graph$. The complexity should be maximal if a task $f_v$ is infeasible for all $M \in \modelAS$, i.e., the WL coloring $\WL^L_v(\graph)$ of $v$ in the graph $\graph$ contains insufficient information to deduce $f_v(\graph)$:%
\begin{definition}
For $\graph, v \in \graphsV$, a task $f$ over $\graphsV$ and an $L$-layer standard MPNN $\modelAS$, define
\[\WLComp_\modelAS(f_v, \graph) = \begin{cases}\infty & \text{if }\WL^L_v(\graph) \nvDash_\graphsV f_v(\graph) \\ 0 & \text{else.} \end{cases}\]
\end{definition}
While $\WLComp$ considers specific tasks $f_v$ and graphs $\graph$, it still inherits a fundamental limitation of iso expressivity: it only distinguishes between possible (complexity 0) and impossible (complexity $\infty$) tasks. This binary characterization fails to capture the varying difficulties observed in practice. Moreover, as discussed in \cref{sec:expressivity}, the WL test assumes lossless information propagation between nodes, contrasting sharply with the lossy message passing observed in trained real-world MPNNs. 

\subsection{Weisfeiler \& Leman Go Lossy}
\label{subsec:lossy}
To extend our complexity measure beyond the binary characterization of the standard WL test, we need to account for the \hlc{binary}{varying difficulty} arising from the \hlc{lossy}{lossy message passing observed in practice}.
This difficulty inherently relies on the graph topology, i.e., the difficulty of propagating a message from one node $u$ to another node $v$ depends on 1) the number of $ L$-length walks connecting $v$ and $u$ and 2) the degrees of the nodes on the walk. This can be formalized as the random walk probability $\normInf_{vu}^L$ from $v$ to $u$ with edge probabilities $\normInf$ and has been connected to the amount of gradient information node $v$ receives from $u$ \cite{jumping_knowledge} and to the oversquashing phenomenon \cite{on_over-squashing}. Correspondingly, $\normInf_{vu}^L$ serves as a measure of difficulty for the simple task of propagating a message from $u$ to $v$.

How can we adapt this measure of difficulty to \emph{arbitrary} tasks?
For this, we propose \ourWL{}, a probabilistic variant of the WL test that models the possibility of message loss.
 This allows us to quantify difficulty even for 
  \begin{wrapfigure}{r}{0.38\textwidth} %
\vspace{-0.2cm}
\centering
\includegraphics[width=0.85\linewidth]{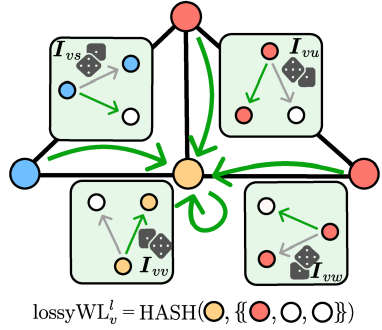} %
\vspace{-0.1cm}
\caption{Update step of \ourWL{} for node $v$. Every message $m^l_{a\to v}$ (green) survives independently with probability $\normInf_{va}$. $\ourWL{}$ models the lossy message propagation observed in real-world MPNNs.}
\vspace{-0.6cm}
\label{fig:pWL}
\end{wrapfigure}
 WL-distinguishable \hlc{hl4}{real-world graphs}. 
 Specifically, a message \( m_{u \to v}^l \) sent from a node \( u \) to its neighbor \( \smash{v \in \neighbor_\graph(u)} \) survives with probability \(\smash{\normInf_{vu}}\) (\cref{fig:pWL}). As a result, the color \(\smash{\ourWL_v^l}\) of a node \( v \) at iteration \( l\) becomes a random variable rather than a deterministic value.

\begin{definition}[\ourWL]
\label{def:lossyWL}
For a given graph $\graph$, let $m_{u \to v}^{l}$ be the (potentially lost) message from $u$ to $v$ at iteration $l$:
\begin{align*}
m_{u \to v}^{l} = Z_{uv}^l \cdot \ourWL^{l-1}_{u} 
\end{align*}
with independent $Z_{uv}^l \sim \mathrm{Bernoulli}(\normInf_{vu})$, indicating whether the message is successfully transmitted.
The node colors of the probabilistic WL tests are then recursively defined as:
\begin{align*}
    \ourWL_v^{0}  &= \features_v \\
\ourWL_v^{l} &= \hash\left(m_{v\to v}^{l}, \ldbrace m_{u \to v}^{l} \mid u \in \neighbor_\graph(v) \rdbrace \right)
\end{align*}
\end{definition}

Similar to how we can transform the WL test into $\WLComp$, we can define our complexity measure, \ours{}, based on our probabilistic \ourWL{} test. To do this, we first extend the \cref{def:fine_grained} of $\vDash$ to account for probabilistic functions such that $\mathbb{P}[\alpha(x) \vDash \beta(x)]$ represents the probability that $\beta(x)$ can be uniquely deduced from the probabilistic output of $\alpha(x)$ (\cref{fig:finegrained_example_prob}).
\begin{definition}
\label{def:vDashProb}
Let $\beta: X \to Z$ be a deterministic function and $\alpha: X \to Y$ be a probabilistic function that can be represented as a deterministic function $\alpha_s$ where $s$ is a seed drawn uniformly at random from a (finite) set $S$.
 For a fixed $x \in X$, define:
\begin{align*}
\mathbb{P}[\alpha(x) \vDash_X \beta(x)] := 
 {\mathbb{P}}_{s \in S}[\forall x' \in X, \, \forall s,t \in S\colon\: \alpha_s(x) = \alpha_{t}(x') \Rightarrow \beta(x) = \beta(x')]
\end{align*}
\end{definition}
With this in place, we define our message passing complexity: intuitively, if there is a high probability of preserving the information needed for $f_v$, the task should have low complexity, and vice versa.
\begin{definition}[\ours{}]
    \label{def:MPC}
    For $\graph, v \in \graphsV$, a function $f$ over $\graphsV$ and an $L$-layer MPNN $\modelAS$, define
    \begin{align*}
        \ours_\modelAS(f_v, \graph) = - \log\mathbb{P}[\ourWL^L_v(\graph) \vDash_\graphsV f_v(\graph)]
    \end{align*}
\end{definition}
\ours{} is based on the probability that under lossy message passing on $\graph$, the output after $L$ layers contains sufficient information to deduce $f_v$. Intuitively, a high \ours{} value indicates that the task $f_v$ requires 1) combining information from many nodes through 2) messages of low probability, such as messages through bottlenecks. Conversely, a low $\ours{}$ value means the task depends on information that remains reliably accessible even under lossy message passing conditions. Note that \ours{} implicitly also depends on $\graphs$, which we omit for simplicity. While we focus here on standard MPNNs $\modelAS$, \ours{} generalizes naturally to a wide range of architectures $\modelA$ by performing \ourWL{} on modified message passing graphs (see \cref{app:mpnn}). 
All in all, \ours{} effectively addresses the discussed limitations of iso expressivity theory, narrowing the gap between theory and practice.

\subsection{\ours{} describes theoretical and practical limitations of MPNNs}
\label{subsec:theory}
Having defined our complexity measure $\ours$, we now demonstrate that it unifies theoretical expressivity with practical GNN limitations in a single framework. We first establish that $\ours$ preserves impossibility results from iso expressivity theory, then show how it captures practical phenomena such as over-squashing and under-reaching. We begin by characterizing when $\ours$ becomes infinite, which precisely corresponds to tasks that are theoretically impossible for a given architecture $\modelA$.\footnote{We defer all proofs to \cref{app:proofs}.} 
\begin{restatable}[Infeasibility]{theorem}{Infeasibility}
    \label{theo:infeasibility}
    The complexity for $\graph,v \in \graphsV$ and function $f$ is
    $\ours_\modelA(f_v, \graph) = \infty$
    if and only if there exist $\graph',w \in \graphsV$ such that $f_v(\graph) \neq f_w(\graph')$ but $\model_v(\graph) = \model_w(\graph')$ for all model instantiations $\model \in \modelA$.
\end{restatable}
This shows that $\ours$ becomes infinite precisely when a model architecture fundamentally cannot distinguish between two different nodes that require different outputs.
We can also globally characterize the functions an architecture can express:
\begin{restatable}{lemma}{recover}
\label{lem:recover}
There exists no model instantiation $\model \in \modelA$ such that $\model_v(\graph) = f_v(\graph)$ for all $\graph,v \in \graphsV$ if and only if there exists $\graph,v \in \graphsV$ with $\ours_\modelA(f_v, \graph) = \infty$. 
\end{restatable}
By choosing an isomorphism test (like the WL test) as task $f$, we can recover all impossibility statements from iso expressivity theory (\cref{lem:iso_subsumes}). Thus, $\ours$ subsumes iso expressivity theory while providing a continuous difficulty measure for possible tasks.
Having established when \ours{} becomes infinite, we now characterize how \ours{} scales with function granularity:  a finer, more discriminative task can never have lower complexity than the coarser task it refines. For example, counting the exact number of cycles in a graph should be more complex than merely detecting whether any cycle exists.
\begin{restatable}[Function refinement]{theorem}{refinement}
    \label{theo:refinement}
    Let $f$ be a function that is more fine-grained than $g$, i.e., $f \vDash_\graphsV g$. Then for any $\graph,v \in \graphsV$:
    $\ours_\modelA(f_v,G) \ge \ours_\modelA(g_v,G)$.
\end{restatable}
Like other complexity measures, MPC satisfies compositionality: solving tasks jointly cannot be more complex than solving them separately, and may be easier when tasks share information.
\begin{restatable}[Task Triangle Inequality]{lemma}{triangle}
\label{lem:task_triangle}
Let $f$ and $g$ be functions, and $\Vert$ denote concatenating function outputs. Then for any $\graph,v \in \graphsV$:
$\ours_\modelA(f_v \Vert g_v, \graph) \le \ours_\modelA(f_v, \graph) + \ours_\modelA(g_v, \graph).$
\end{restatable}
However, \ours{} connects not only to properties of classical complexity and expressivity theory but also captures real-world limitations of MPNNs such as over-squashing and under-reaching. These phenomena, extensively studied as significant constraints on MPNN performance, are overlooked by traditional expressivity theory. We first relate \ours{} to over-squashing by showing that it is lower bounded in terms of the $L$-step random walk probability from $v$ to $u$, $(\normInf^L)_{vu}$, a quantity that also motivated our design of lossyWL in \cref{subsec:lossy}.
\begin{lemma}[Informal version of \cref{lem:lower_bound_formal}]
\label{lem:lower_bound}
Consider a task $f$ and  $\graph,v \in \graphsV$ where $f_v$ "depends on" information $\features_u$ from a node $u$. Then
\begin{equation*}
\ours_\modelA(f_v, \graph) \ge  - \log \left ( (\normInf^L)_{vu} \right ).
\end{equation*}
\end{lemma}
 A lower probability $(\normInf^L)_{vu}$ indicates that node $v$ receives less gradient signal from $u$, making it more susceptible to over-squashing \cite{on_over-squashing, jumping_knowledge}. Our complexity measure captures this: when $(\normInf^L)_{vu}$ is small, tasks requiring information flow from $u$ to $v$ have high complexity.
As a simple special case, we consider under-reaching, which occurs when the number of MPNN layers is less than the graph diameter, preventing nodes from receiving information from distant parts of the graph \cite{under_reaching,oversquashing_1}:
\begin{corollary}
\label{cor:under_reaching}
Consider a task $f_v$ and a graph $\graph \in \graphs$ where $f_v$ "depends on" information $\features_u$ from a node $u$ outside of the receptive field, i.e. $d_\graphT(u,v) > L$. Then
$\ours_\modelA(f_v, \graph) = \infty.$
\end{corollary}

These results show that \ours{} captures both theoretical expressivity and practical limitations of GNNs, providing a more unified framework that bridges both perspectives.

\vspace{-0.2cm}
\section{MPC in Practice: Explaining Empirical MPNN Behavior}
\vspace{-0.12cm}
\label{sec:evaluation}
 We now demonstrate that \ours{} quantitatively explains \emph{empirical} MPNN behavior. Through analysis of fundamental graph tasks—retaining information, propagating information, and extracting topological features—we show that:
 1) \ours{} enables the derivation of meaningful task-specific complexity bounds; 2) trends in MPC complexity correlate strongly with empirical performance, better reflecting the practical behavior of MPNNs than iso expressivity theory; 3) optimal architectures vary by task, with success determined by minimizing task-specific \ours{} complexity through appropriate inductive biases, not maximizing expressivity.
\textbf{Experimental Setup:} 
We verify this for a wide range of architectures, including an MLP baseline (message passing on an empty graph), standard models like GCN \cite{gcn}, GIN \cite{xu}, GraphSage \cite{sage}, and a GCN with a virtual node, GCN-VN \cite{vn}, as well as higher-order models like  GSN \cite{gsn} which incorporates substructure-information as node-feature, FragNet \cite{fragNet} which builds a higher-level graph of fragments and the topological-inspired model CIN \cite{cw}. We evaluate on random $r$-regular graphs $\graphs$ and show that the results transfer to graphs from the ZINC dataset \cite{ZINC} and the long-range graph benchmark \cite{lrgb}. We provide theoretical complexity bounds and Monte-Carlo simulated complexity values. We compare \ours{} to the WL-based WLC baseline rather than directly to other expressivity measures, as they have fundamental limitations for task-specific difficulty analysis: they typically provide only global, task-agnostic architecture rankings \cite{substructure_wl,biconnectivity} or impose restrictive assumptions incompatible with our tasks \cite{mixing} (see 
\cref{sec:related_work}).

\textbf{Retaining information}\quad
 First, we evaluate the task 
$f_v(\graph) = \features_v$,
 which tests a model’s ability to retain its initial node features. This task is fundamentally important across all domains with informative node features (e.g., atom type in molecules). Although seemingly simple, it is closely linked to the well-studied \emph{over-smoothing} \cite{oversmoothing} phenomenon, where deeper GNNs lose distinguishability between nodes as their representations converge, making it difficult to recover the original node features \cite{oversmoothing2}.
Our complexity measure captures this difficulty as \ours{} increases at least linearly with depth $L$:
\begin{restatable}{lemma}{retaining}
\label{lem:retaining}
Assume the degree $r \ge 2$.
Then the expected \ours{} complexity for this task  $\mathbb{E}_{\graph, v \sim \graphsV}[\ours_\modelA(f_v, G)]$ grows at least linearly with $L$, i.e., is in $\Omega(L)$, for all MPNNs $\modelA$ .
\end{restatable}
\begin{wrapfigure}{r}{0.48\linewidth}
    \centering
    \vspace{-0.3cm} \includegraphics[width=0.8\linewidth]{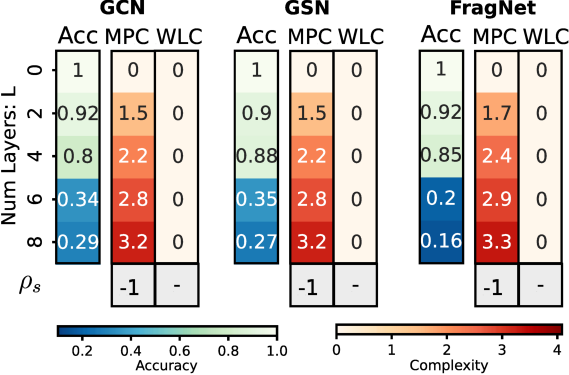}
    \vspace{-0.2cm}
    \caption{Test accuracy for retaining initial node features compared with complexity measures \ours{} and \WLComp{}. Simulated \ours{} (in contrast to WL-based \WLComp{}) shows perfect negative Spearman correlation ($\rho_s = -1$) with accuracy, capturing increasing difficulty with depth (over-smoothing). Complete results in \cref{fig:keep_heatmap_full}.}
    \label{fig:keep_heatmap}
    \vspace{-0.2cm}
\end{wrapfigure}

For empirical validation, we train all model architectures on 2000 randomly generated $3$-regular graphs with varying numbers of layers $L$. As shown in \cref{fig:keep_heatmap,fig:keep_heatmap_full}, the complexity measure based on iso expressivity theory, \WLComp{}, assigns constant zero complexity regardless of depth $L$, indicating only theoretical solvability.
In contrast, \ours{} shows perfect negative Spearman correlation with accuracy for most architectures, capturing the increasing difficulty with $L$. Only CIN and GraphSage maintain perfect accuracy throughout, due to explicit residual connection optimization—an implementation choice our framework abstracts from.
These results demonstrate that, unlike binary expressivity measures, \ours{} accurately quantifies the progressive difficulty of this task for most architectures, capturing important real-world limitations beyond theoretical impossibility statements.

\begin{figure}[tb]
    \centering
    \begin{minipage}[t]{0.42\linewidth}
        \centering
        \includegraphics[width=\linewidth]{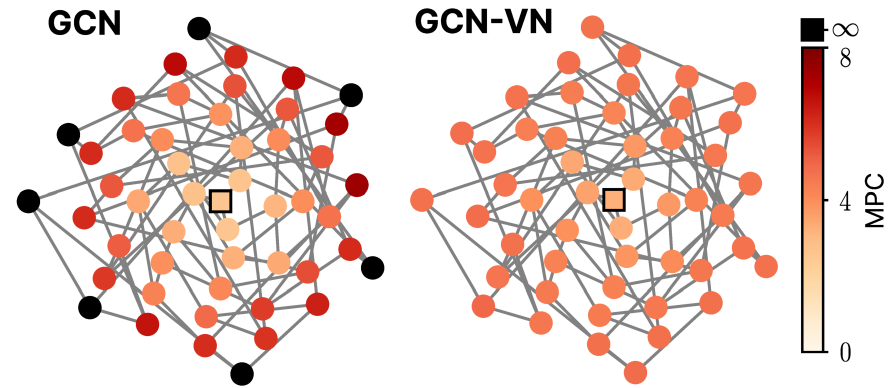}
        \vspace{-0.6cm}
        \caption{Simulated \ours{} complexities for propagating features from source nodes $u$ (colored by \ours{}) to target node $v$ (square). Despite identical iso expressivity, \ours{} reveals the significant advantage virtual nodes offer for long-range dependencies.}
        \label{fig:vn_example}
    \end{minipage}%
    \hfill
    \begin{minipage}[t]{0.55\linewidth}
        \centering
        \includegraphics[width=\linewidth]{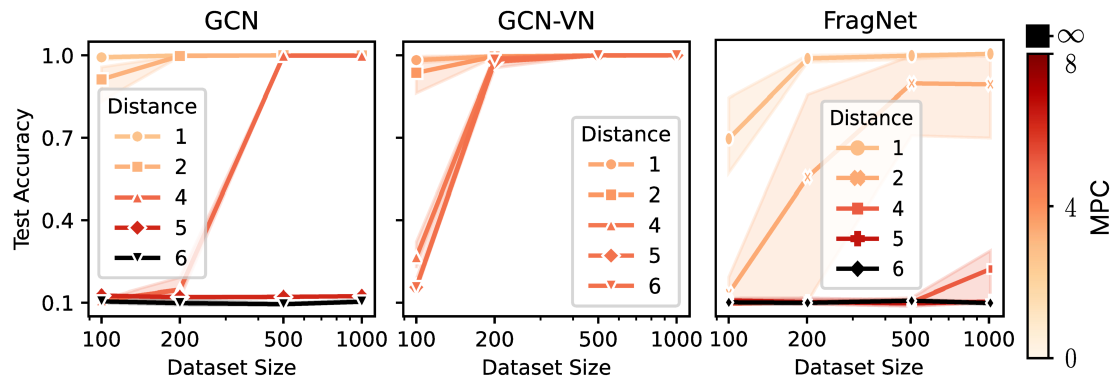}
        \vspace{-0.6cm}
        \caption{Test accuracy vs. training data size for the propagation task $f_v(\graph) = \features_u$ for different distances $D$. Colors indicate average simulated \ours{} for each distance. Higher MPC values reflect greater task difficulty, evidenced by increased sample complexity. All results in \cref{fig:transfer_full}; for real-world graphs from lrgb in \cref{fig:peptides_line}.}
        \label{fig:transfer_selected}
    \end{minipage}
    \vspace{-0.35cm}
\end{figure}

\textbf{Propagating Information}\quad  
We next analyze the task $f_v(\graph) = \features_u$, where $u$ is a specially marked source node at distance $D = d_\graph(u,v)$ from target node $v$. This task directly tests a model's ability to propagate information in relation to the distance $D$, exposing practical limitations like over-squashing and under-reaching that classical expressivity theory overlooks. 
\begin{restatable}{lemma}{transfer}
\label{lem:comp_transfer}
Assume $L$ is the minimum depth required to solve this task with $\modelA$. Then the expected complexity $\mathbb{E}_{\graph,v \sim \graphsV}[\ours_\modelA(f_v,G)]$ is  $ \le 2\log(n)$ for GCN-VN, while for standard MPNNs it is $\ge D\log(r)$ provided $n$ is sufficiently large given $D$.
\end{restatable}
The bounds reveal that \ours{} captures two critical insights missed by classical expressivity: First, \ours{} increases with distance $D$ for most MPNNs, explaining why they struggle with long-range dependencies despite theoretical learnability. Second, a virtual node fundamentally changes complexity scaling from $O(D)$ to $O(\log n)$, explaining its empirical advantage despite unchanged expressivity (see \cref{fig:vn_example}). Moreover, \ours{} also captures under-reaching: by \cref{cor:under_reaching}, complexity becomes 
infinite when $d_\graphT(u,v) > L$.

Experiments for models with $L=5$ layers validate these predictions:  \ours{} correlates negatively with accuracy (\cref{fig:transfer_full_heatmap}) and captures the increased sample complexity with distance (\cref{fig:transfer_selected,fig:transfer_full}). Crucially, all models except GCN-VN fail at $D=5$, showing that high \ours{} complexity indicates practical limitations even before the theoretical impossibility at $D>5$. Importantly, these findings—strong MPC-performance correlation and practical failure before theoretical limits—persist in real-world graphs from the long-range graph benchmark (\cref{fig:peptides_line}).

In summary, these results demonstrate why \ours{} offers a more complete and practical understanding of MPNN capabilities than classical expressivity theory. Unlike binary expressivity tests, \ours{} explains why even highly expressive architectures, such as GSN, FragNet, and CIN, struggle with fundamental tasks like retaining or propagating node information—empirically validating our critique in \cref{sec:expressivity} that iso expressivity poorly captures practical capabilities. Furthermore, \ours{} accounts for the success of architectural choices like virtual nodes, which consistently improve performance on real-world long-range tasks \cite{vn_empirical} despite unchanged iso expressivity \cite{fragNet}.

\textbf{Topological Feature Extraction} \quad
Our final experiment examines how models extract topological features through cycle detection. We consider random $r$-regular graphs with unique node labels $\features_v \in \{1,\ldots,n\}$, modified to contain a cycle of size $s$ that includes a designated node $v$. The task $f_v$	is to identify the labels of all nodes in this cycle, jointly testing the model’s ability to detect cycles and propagate information across them. We can derive complexity bounds for all considered MPNNs:%
\begin{lemma}[Informal version of \cref{lem:ring_prop_formal}]
\label{lem:ring_bounds}
Assume $L$ is the minimal depth required to solve this task with architecture $\modelA$ and that there is only a single cycle in the $\lceil s/2 \rceil$-hop neighborhood of $v$. Then, the expected \ours{} complexities are: 
\[\text{For CIN \& FragNet:}\:\: O(\log(sr)) \quad \quad \text{For GSN:} \:\: \leq \lceil s/2 \rceil \log(r+1) \quad\quad \text{For all others:}\:\: \geq s \log(r)\]
\end{lemma}
\begin{wrapfigure}{r}{0.57\textwidth} %
\centering
\vspace{-0.4cm}
\includegraphics[width=0.9\linewidth]{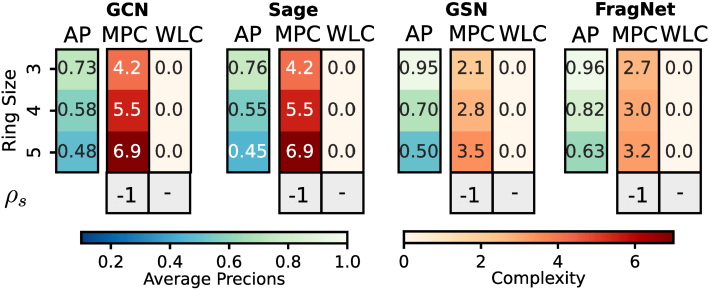} %
\vspace{-0.09cm}
\caption{Average precision (AP) for the ring task compared with complexity values \ours{} and \WLComp{} (\ours{} values for SAGE \& GCN are lower bounds).  \ours{} (in contrast to \WLComp{}) is negatively Spearman correlated with AP and captures the advantage of cycle-oriented inductive biases in GSN and FragNet. See also \cref{fig:ring_heatmap_full,fig:ZINC_line}.}
\vspace{-0.7cm}
\label{fig:ring_heatmap}
\end{wrapfigure}
These bounds reveal that \ours{} captures crucial architectural differences missed by iso expressivity: standard MPNNs have complexity scaling linearly with cycle size, while CIN and FragNet achieve logarithmic scaling, capturing their cycle-oriented inductive biases. Experiments on random 4-regular graphs confirm these theoretical predictions, with trends in \ours{} complexity matching empirical performance (see \cref{fig:ring_heatmap,fig:ring_heatmap_full}; in relation to dataset size: \cref{fig:ring_line_full}).

Because of the unique node labels, all (at least WL expressive) architectures are theoretically able to solve this task, yet performance varies dramatically. While iso expressivity theory cannot capture these differences, \ours{} provides a clear explanation: GSN, CIN and FragNet's architectural bias towards cycles reduces their message-passing complexity for this task, providing them with a performance advantage. This insight extends to real-world graphs from the ZINC molecular regression benchmark \cite{ZINC} (\cref{fig:ring_heatmap}), where the identification of molecular rings is crucial. As shown in \cref{tab:ring_zinc_comparison}, architectures with lower ring detection complexity consistently achieve better performance on ZINC on both the synthetic and the real task. The superior performance of higher-order MPNNs like GSN, FragNet or CIN cannot be explained by their greater iso expressivity—as discussed in \cref{sec:expressivity}, almost all graphs in ZINC are already WL-distinguishable—but by their cycle-oriented inductive biases that reduce the complexity of learning molecular structures.

Overall, our results demonstrate that trends in \ours{} consistently correlate with empirical performance across all tested tasks, offering a more fine-grained, complete, and practical understanding of MPNN capabilities than iso expressivity theory's idealized binary characterization. Crucially, we find that no single architecture performs best across all tasks---performance depends not on iso expressivity but on the alignment between a model’s inductive biases and the task requirements. While the global task-agnostic architecture rankings of most existing expressivity measures fail to capture this variation, \ours{} quantifies the alignment between architecture and task---explaining why virtual nodes excel at long-range tasks or why cycle-aware designs help for ring-transfer. This provides task-specific guidance for architectural design that pure expressivity theory cannot offer.

\vspace{-0.1cm}
\section{Limitations \& Future Work}
\vspace{-0.1cm}
\label{sec:limitations}
While we have demonstrated that our complexity measure closely aligns with empirical performance for many tasks, it abstracts away from the parametrization and implementation of the $\upd$ and $\AGG$ functions and the optimization process as discussed in \cref{sec:evaluation}. Additionally, our MPNN framework does not include models with attention. However, we repeat all our experiments with the most common attention-based MPNNs in \cref{app:experimental_details}, showing similar empirical trends to the MPNNs that we consider.
Secondly, exact target functions in real-world applications are usually unknown. Still, we can analyze two types of proxy tasks: fundamental capabilities required across all domains (like retaining information) and domain-specific operations (like cycle detection). As shown in \cref{tab:ring_zinc_comparison}, the insights from these proxy tasks can translate to real-world performance.
Third, exact complexity values can be difficult to compute for sophisticated tasks. However, as our experiments demonstrate, theoretical bounds can provide valuable practical insights.
Finally, the complexity values should not be interpreted in isolation to determine whether an architecture can solve a given task, as they only characterize the \emph{message-passing} complexity of a task. Instead, they are most useful for identifying trends and comparing architectures across a single dimension of freedom.

These limitations point to promising extensions of the MPC framework. \ours{} could be extended to incorporate additional sources of practical difficulty, such as feature noise (\cref{subsec:noise}), providing a more complete view of empirical task complexity. Furthermore, the framework could be modified to capture aggregation and update-specific effects by incorporating non-uniform or learnable message weights $Z_{uv}$ (as in attention-based models), enabling analysis of how different update functions affect message passing complexity.

More broadly, we hope that \ours{} will enable more principled architectural design: By quantifying fine-grained task complexities, \ours{} can reveal \emph{practical} limitations of current architectures and guide the development of models with low complexity \emph{for specific capabilities} relevant to target domains. As a concrete example, consider substructure identification. Our analysis in \cref{sec:evaluation} shows that all considered MPNNs without additional pre-processing steps exhibit very high complexity for this task, even when it is theoretically solvable. This suggests that architectural modifications alone may be insufficient, pointing toward studying expressive preprocessing steps or positional encodings ---as successfully employed by FragNet \cite{fragNet} and CIN \cite{cw}---  that directly reduce task complexity. More generally, this exemplifies how MPC can shift architectural design from maximizing universal expressivity toward strategically minimizing complexity for domain-specific requirements.

\section{Related Work}
\label{sec:related_work}

\textbf{Iso-expressivity frameworks.} The predominant approach for analyzing GNNs compares their ability to distinguish non-isomorphic graphs relative to the WL test and its extensions \cite{morris,xu,overview_wl}. This has motivated extensive research on developing higher-order architectures that surpass standard MPNNs in expressivity \cite{overview_wl,cw,cinpp,fragNet}. However, as discussed in \cref{sec:expressivity}, iso-expressivity theory relies on idealized assumptions and provides only binary characterizations, limiting its practical relevance.

\textbf{Alternative expressivity characterizations.} Beyond iso-expressivity, several frameworks theoretically characterize MPNN capabilities through specific tasks. \citet{substructure_wl} rank architectures by the set 
of substructures they can recognize, while \citet{biconnectivity} compare models through their ability to solve graph biconnectivity. \citet{spectral} analyze spectral MPNNs through their ability to learn polynomial filters. Logic-based approaches \cite{under_reaching,logic} characterize learnable functions through fragments of first-order logic, providing more nuanced insights by considering, for example, the effects of different activation functions \cite{logic2}.  However, these expressivity characterizations share key limitations with iso expressivity: they remain binary (can/cannot solve) and assume lossless information propagation, limiting their insights for real-world MPNNs.

\textbf{Continuous graph distances.}  To move beyond binary expressivity, some works have proposed continuous graph similarity measures, including tree-based \cite{tree_distance}, graphon-based \cite{graphon_distance}, and Wasserstein-based distances \cite{wasserstein_distance}. However, these metrics are task-agnostic and architecture-independent, limiting their relevance for explaining practical GNN performance.

\textbf{Practical GNN limitations.} A parallel line of research studies individual \emph{practical} limitations of GNNs, such as over-smoothing \cite{oversmoothing}, under-reaching \cite{under_reaching}, and over-squashing \cite{oversquashing_1,on_over-squashing,oversquashing_curvature}. Most relevant to our approach is the work by \citet{mixing}, which derives expressivity limitations from over-squashing theory. They show tasks become impossible when the required "mixing" between nodes (measured via the Hessian) exceeds what MPNNs can generate. Like MPC, this can capture practical impossibilities arising from over-squashing. However, their framework considers only pairwise interactions with restrictive assumptions on the task (twice differentiable tasks, not dependent on graph topology).

While these works study individual theoretical or practical limitations of GNNs, \ours{} is the only framework that captures both theoretical expressivity and practical limitations while allowing analysis of arbitrary tasks.

\section{Conclusion}
\label{sec:conclusion}
We show that classical expressivity theory cannot explain MPNN performance in real-world settings. To narrow the gap between theory and practice, we propose \ours{}: a continuous complexity measure that quantifies the message-passing difficulty of tasks for different architectures. By building upon a novel probabilistic variant of the WL test, \ours{} retains all impossibility results from iso expressivity theory while capturing practical limitations like over-squashing and under-reaching. Our extensive validation on fundamental MPNN tasks reveals that trends in \ours{} complexity correlate with empirical performance, explaining phenomena that iso expressivity theory cannot address. Notably, our analysis indicates that the success of (higher-order) MPNNs often stems from low task complexity rather than increased iso expressivity. This perspective shifts focus from maximizing expressivity to minimizing task-specific \ours{} complexity, providing clear directions for architectural innovation.

\section*{Acknowledgments}
We would like to thank Nicholas Gao and Aman Saxena for proofreading this manuscript, and the anonymous reviewers for their constructive comments. This research is supported by the Bavarian Ministry of Economic Affairs, Regional Development and Energy with funds from the Hightech Agenda Bayern.

\bibliography{references/references}

\appendix

\newpage

\section{Extended Related Work}
\label{sec:extended_related_work}
Existing efforts to address the limitations of iso expressivity theory typically target only one limitation, focusing on either overcoming its task-agnostic nature, the unrealistic assumption of lossless information propagation, or its binary characterization of expressivity. 

\paragraph{Task-Specific expressivity}
While iso expressivity theory is task-agnostic, recent works have proposed studying specific graph-related tasks to get a more fine-grained hierarchy of the expressivity of higher-order MPNNs (beyond the WL test). For example, expressivity has been analyzed in the context of substructure recognition \cite{substructure_wl} and graph biconnectivity \cite{biconnectivity}.  

A separate line of work characterizes the node-level functions computable by MPNNs in terms of fragments of first-order logic \cite{logic}: \citet{under_reaching} showed that standard MPNNs with sum aggregation and ReLU activations can uniformly express graded modal logic. Subsequent work analyzed the effect of different activation functions \cite{logic2}, and more recently, these logical characterizations were extended to various higher-order MPNNs \cite{logic_all}.

However, all these works only provide \emph{existence} results: they show whether there exists some model instance of an architecture that can solve a task or not. Such results neglect practical limitations like information loss (seem in phenomenons like over-squashing or over-smoothing) occurring in \emph{trained real-world models}.
Moreover, the characterizations are inherently binary (solvable or not), offering limited insight into the practical difficulty. 

\paragraph{Accounting for Lossy Information Propagation.} Some works explicitly account for the lossy information propagation inherent in practical GNNs. Negative results have been derived by considering limited hidden dimensions \cite{communication_capacity} or by analyzing the maximal "mixing" of node representations \cite{mixing}. However, these approaches are restricted to specific model architectures and provide binary results, rather than a nuanced understanding of how lossy propagation impacts practical performance in arbitrary tasks. 

\paragraph{Beyond Binary Expressivity.} To address the binary nature of classical expressivity theory, some works define distances between graphs to represent the difficulty of distinguishing them. Examples include tree-based distances \cite{tree_distance}, graphon-based distances \cite{graphon_distance}, and Wasserstein-based distances \cite{wasserstein_distance}. While these metrics provide a continuous measure, they remain task-agnostic and are independent of model architecture.

\paragraph{Restricted Graph Families.} Classical expressivity theory evaluates GNNs on the set of all possible graphs, which may not align with the restricted graph families encountered in real-world applications. Recent works examine expressivity within specific graph families, such as planar graphs \cite{planar} and outer-planar graphs \cite{outer_planar}, providing more relevant insights for certain domains.

\paragraph{Higher-order MPNNs and variants of the WL Test.} 
\begin{wraptable}{r}{3.8cm}
\vspace{-0.65cm}
    \centering
    \sisetup{detect-all}
    \caption{Fraction of graphs with unique WL hashes (ignoring isomorphic graphs). Similar to \citet{wl_enough}.}
    \vskip 0.05in
    \resizebox{3.8cm}{!}{\begin{tabular}{
  @{}
  l
  r
  @{}
}
\toprule
\textbf{Dataset} & \textbf{Unique WL} \\
\midrule
Reddit-Binary & 100\% \\
Peptides & 100\% \\
Mutag & 100\% \\
Enzymes & 100\% \\
Protein-dataset & 100\% \\
ZINC-subset & 100\% \\
ZINC-full & $>99.99$\% \\

\bottomrule
\end{tabular}
}
    \label{tab:wl}
\vspace{-0.7cm}
\end{wraptable}
A plethora of higher-order MPNNs \cite{cw,simplicial,fragNet,morris} have been developed to surpass the WL test in iso expressivity. To quantify and compare their expressivity 
corresponding (higher-order) variants of the WL test \cite{morris,cw,fragNet,neighbor_wl} have been developed. 
 However, we show in \cref{tab:wl} similar to \citet{wl_enough} that the standard WL test already suffices to distinguish almost all graphs in standard benchmarks. Hence, expressivity theory statements that only focus on expressivity beyond the WL test cannot offer explanations for performance differences of models that are at least WL expressive on these popular benchmarks.

Many of these higher-order methods, along with an overview of WL variants, are comprehensively surveyed in \citet{overview_wl}.

\paragraph{Real world limitations of GNNs}
Several fundamental limitations of GNNs have been identified in practical applications. Under-reaching \cite{under_reaching} occurs when nodes cannot access information from distant parts of the graph due to insufficient layer depth, effectively limiting the receptive field of each node. Over-squashing \cite{oversquashing_1,oversquashing_curvature,on_over-squashing} describes how graph topology can create bottlenecks in message passing, preventing effective information flow between nodes even when they are theoretically within each other's receptive field. This is distinct from over-smoothing \cite{oversmoothing}, where increasing the number of layers causes node representations to become indistinguishable as all nodes converge to the same representation. These limitations highlight a crucial gap between theoretical expressivity and practical capabilities of GNNs.

\paragraph{Practical power of standard GNNs}
A few other works highlight a disconnect between theoretical expressivity limitations and the practical capabilities of standard GNNs. While expressivity theory establishes that standard GNNs cannot count any non-trivial substructure \cite{substructure_counting}, recent studies reveal important exceptions. \citet{gnns_can_count} identify specific conditions on the graph family $\graphs$ under which standard GNNs can count substructures and demonstrate that many real-world datasets satisfy these conditions for important substructures like small cycles. Similarly, \citet{counting_random} showed that adding randomized node features enables standard GNNs to count small cycles. 

\paragraph{Task-Model Alignment.} Beyond expressivity, our work aligns with efforts to quantify the compatibility between a model and its target task. For instance, \citet{alignment} define on a higher level the concept of algorithmic alignment between a general neural network and an algorithm, showing that higher alignment leads to lower sample complexity. Similarly, our proposed complexity measure can be interpreted as quantifying the alignment between a specific graph task and the message passing steps needed to solve it for a given MPNN architecture.

In summary, while existing works address individual limitations of classical expressivity theory, none provide a unified framework addressing all limitations.

\section{Extending \ours{}: More Architectures, Graph-Level Functions, and Feature Noise}
In the following, we show how \ours{} can be extended to a wide range of architectures beyond standard MPNNs and to graph-level functions. Additionally, we sketch a potential extension to include additional sources of complexity, such as feature noise. Furthermore, we sketch how one could easily adapt the framework to operate on temporal graphs.

\subsection{General Message Passing Framework}
\label{app:mpnn}
In the following, we provide a general message passing framework for \ours{} that captures a wide range of existing MPNN architectures $\modelA$.

While standard MPNNs $\modelAS$ perform message passing directly on the input graph $\graph$, other architectures operate on a transformed message passing graph $\graphT = t(\graph)$:
\begin{align*}
      m_{w \to v}^l := \begin{cases} \msg_0^l(h_w^{l-1}) & \text{if}\:\: w=v \\ \msg_1^l(h_w^{l-1}) & \text{else} \end{cases} \quad \text{and}\quad h_v^l := \upd^l (\AGG \left ( \ldbrace  m_{w \to v}^l \mid w \in \neighbor_\graphT(v) \cup \{v\} \rdbrace \right ) )
 \end{align*}
The transformation $t$ defines the MP graph structure $\graphT$ by determining which nodes exchange messages. While standard MPNNs used the input graph $\graph$ directly as $\graphT$, recent architectures introduce modifications like virtual nodes \cite{vn}, rewired edges \cite{oversquashing_curvature}, or higher-order graphs \cite{fragNet}.  We assume $t$ preserves original nodes $\vertices$ while potentially adding nodes or adding/modifying edges. A general MPNN architecture $\modelA$ is defined through the transformation $t$ and the number of layers $L$, characterizing which nodes exchange information. It abstracts away from the specific choice of aggregation method and update function, which is a sensible simplification since many recent models treat the choice of aggregation method as a hyperparameter and use MLPs as update functions (providing universal function approximation capabilities).

This framework captures many existing MPNNs (potentially with some simplifying assumptions):

\textbf{Standard MPNNs}\quad
Standard architectures like GCN \cite{gcn}, GraphSage \cite{sage}, GIN \cite{xu} perform message passing directly on the input graph, i.e., $\graphT = t(\graph) = \graph$. They differ in aggregation method (mean, sum, degree-normalized mean). Some architectures, like GCN, restrict the possible choices of $\msg$ and $\upd$ function and do not distinguish self-loops from normal edges, i.e., $\msg_0 = \msg_1$. Note that, as discussed above, our MPNN framework abstracts away these architectural choices.

\textbf{Virtual nodes, rewiring and additional features}\quad
Modifications to the input graph such as additional virtual nodes $\cite{vn}$, edge rewiring \cite{oversquashing_curvature}, or additional node features (e.g., including substructure information like in GSN \cite{gsn})  can be directly modeled in our framework by choosing the appropriate transformation function $t$.

\textbf{Higher-order MPNNs}\quad
Higher-order MPNNs that augment the input graph with higher-level structures are also representable.
FragNet \cite{fragNet} builds an additional higher-level graph of fragments (for an arbitrary fragmentation scheme). We will focus here on the FragNet model with a fragmentation scheme identifying rings without edge representations. We model this by adding additional fragment nodes connected to all constituent input nodes, with labels corresponding to fragment types. The CIN \cite{cw} model follows a similar approach by having representations for all original nodes, edges, and CW-cells (rings). Upper messages are messages from nodes to their corresponding edge representation, and from edges to the corresponding ring representation (if any). Boundary messages are messages between nodes sharing an edge and between edges that are both in the same ring. The initial representation of CW cells is an aggregation of the features of their constituent nodes.
We represent edges and CW cells (rings) by additional nodes in $\graphT$. The initial feature of a ring node is the aggregation of all node features of the ring. We have edges between: neighboring original nodes, original nodes and their edge nodes, edge nodes within the same ring, and edge nodes and their rings (if any). Note that we cannot represent in our framework that boundary messages also contain information from the corresponding upper neighborhood. Importantly, both fragNet and CIN use different update functions for different types of messages. This too can not be modeled in our formulation of $\modelA$; instead, we assume just one update function. While one could, in theory, extend our framework to include different kinds of message updates, we choose against this as our existing framework, with this simplification, already captures the models' empirical performance well.

\textbf{3D GNNs: DimeNet}\quad
Next, we sketch how our framework can model GNNs incorporating 3D information, exemplified by DimeNet \cite{dimeNet}. This architecture maintains representations for all directed edges and incorporates angular information. In our framework, this can be captured by using the directed line graph $\graphT = (\tilde{V}, \tilde{E})$ of the original graph $G$ as the message passing graph where $\tilde{V} = E$ (the original edges become nodes) and for any $e_1 = (v_j, v_k), e_2 = (v_k, v_i) \in E$, we have $(e_1, e_2) \in \tilde{E}$. The angular and distance features of DimeNet can then be incorporated as edge/node features in this transformed message passing graph.

Given this general  message passing framework, we can naturally extend MPC from standard MPNNs $\modelAS$ to general MPNNs $\modelA$ by applying $\ourWL$ to the transformed message passing graph $\graphT = t(\graph)$ rather than the input graph:
\begin{definition}[\ours{}]
    For $\graph, v \in \graphsV$, a function $f$ over $\graphsV$ and a $L$-layer MPNN $\modelA$, define
    \begin{align*}
        \ours_\modelA(f_v, \graph) = - \log\mathbb{P}[\ourWL^L_v(t(\graph)) \vDash_\graphsV f_v(\graph)].
    \end{align*}
\end{definition}

Analogously, we can extend our baseline $\WLComp$ to general MPNNs $\modelA$:

\begin{definition}
For $\graph, v \in \graphsV$, a function $f$ over $\graphsV$ and a $L$-layer MPNN $\modelA$, define
\[\WLComp_\modelA(f_v, \graph) = \begin{cases}\infty & \text{if }\WL^L_v(t(\graph)) \nvDash_\graphsV f_v(\graph) \\ 0 & \text{else} \end{cases}\]
\end{definition}

\subsection{Graph-Level Tasks}
\label{subsec:graph-level}
Additionally, we can generalize \ours{} to graph-level tasks $f:\graphs \to \mathbb{R}^k$. A graph-level output of an MPNN architecture $\modelA$ is learned by aggregating all final node representations $h_v^L$ and transforming the aggregated result. This can be represented in our MPNN framework by having an additional readout node $v_\text{global}$ that receives messages from all other nodes only in the $L+1$-th layer. \ours{} is then defined over the graph, node pairs $\{ (\graph, v_\text{global}) \mid \graph \in \graphs \}$.

\subsection{Additional Sources of Complexity: Feature Noise}
\label{subsec:noise}
While \ours{} is designed to isolate the complexity arising from the message passing topology, it can also be extended to include additional sources of complexity arising in practice, such as feature noise, thereby providing a more unified practical understanding of task-difficulty. We sketch here such a possible extension:

Let $\features_v$ be the true feature of node $v$, and let $\tilde{\features}_v$
 be the observed noisy feature according to some noise distribution $\tilde{\features}_v \sim N_v(\features_v)$. Our probabilistic formulation naturally accommodates this additional source of difficulty by modifying the initialization step in \cref{def:lossyWL} to incorporate noise:
 \[ \ourWL_v^0 = \tilde{\features}_v \quad \text{with} \quad \tilde{\features}_v \sim N_v(\features_v).\]

 Notice that this general model of feature noise subsumes missing features as a special case. All in all, this demonstrates a promising direction for incorporating additional practically relevant sources of complexity within a unified framework in future work.

\subsection{Temporal Graphs}
\ours{} can be naturally extended to temporal graphs as well. Consider snapshot graphs $\graph^\tau$ at time point $\tau$, as defined in \citet{temporal}. The complexity for a task $f_v$ then becomes time dependent:
\[
\ours_\modelA^\tau(f_v, G) := - \log \mathbb{P}[\ourWL_v^L(t(\graph^\tau)) \vDash_\graphsV f_v(G)].
\]
This extension opens up interesting directions for future work, such as analyzing how the complexity of specific tasks evolves over time as the underlying graph changes.

\section{Proofs and Extended Theorems}
\label{app:proofs}
In this section, we present all proofs and extended/formal versions of the theorems in the main body. 

\textbf{Extended Notation}\quad Let $\circ$ denote function composition. Additionally, let $\alpha^{-1}$ be the preimage of a function $\alpha: X \to Y$, i.e., $\alpha^{-1}(y) = \{ x \in X \mid \alpha(x) = y \}$. If $X$ is clear from the context, we will simply write $\alpha \vDash \beta$ instead of $\alpha \vDash_X \beta$. We define $\WL_v := \WL^L_v$ and $\ourWL_v := \ourWL_v^L$, i.e., we omit the layer superscript for the final layer $L$. Additionally, we say a message $m_{u \to v}^l$ in the $\ourWL$ test \emph{is lost}, if $Z_{uv}^l = 0$, otherwise we say it was successful. Let $\model_v(\graph)$ denote the final output $h_v^L$ of a model at node $v$. Additionally, we will say a set $\mathcal{S} \subseteq 2^U$ of sets over an universe $U$ is \emph{upward-closed} if $A \in \mathcal{S}$ implies that any $B \subseteq U$ with $A \subseteq B$ is also in $\mathcal{S}$, i.e., $B \in \mathcal{S}$. $\mathbf{1}[\text{condition}]$ denotes the indicator function defined as $1$ if the condition is true and $0$ else. We use the terms "ring" and "cycle" interchangeably. 

We will often represent a probabilistic function $\tau$ with domain $X$ and finitely many possible probabilistic outputs in $Y$ (each with rational probability)  as a deterministic function $\tau(x,s)$ or $\tau_s(x)$ with a seed drawn uniformly at random from a finite set of seeds $S$. We will sometimes (slightly abuse notation and) write $\tau(x,s) \vDash_{X \times S} \beta(x)$ for a non-probabilistic function $\beta: X \to Z$ if
\[\forall x' \in X. \forall s' \in S: \tau(x,s) = \tau(x',s') \implies \beta(x) = \beta(x'), \]
i.e., if one can deduce $\beta(x)$ from the output of $\tau(x,s)$. 
With this notation, we can nicely write \cref{def:vDashProb} as
\[\mathbb{P}[\tau(x) \vDash_X \beta(x)] := \mathbb{P}_{s \in S}[\tau(x,s) \vDash_{X\times S} \beta(x)].  \]

Additionally, we will define the concept of necessary and sufficient messages that we will use in several proofs.

\begin{definition}
\label{def:sufficient}
We call a set $\sS$ of messages (or rather message identifiers $(a,b,l)$) sufficient for a function $f_v$ on a graph $\graph$, if it is possible to deduce $f_v$ from $\ourWL_v$ given that the messages in $\sS$ were successful, or more formally:
\[ \bigwedge_{(a,b,l) \in \sS} Z_{ab}^l = 1 \implies \ourWL_v(t(\graph)) \vDash_\graphsV f_v(\graph).\]
\end{definition}

\begin{definition}
\label{def:necessary}
We call a set $\sS$ of messages necessary for a function $f_v$ on a graph $\graph$, if it not possible to deduce $f_v$ from $\ourWL_v$ if any of the messages in $\sS$ were lost, or more formally:
\[ \exists {(a,b,l) \in \sS}: Z_{ab}^l = 0 \implies \ourWL_v(t(\graph)) \nvDash_\graphsV f_v(\graph).\]
\end{definition}

Lastly, we define the probability of a set $\sS$ of messages as the probability that all messages were successful:
\begin{definition}
    Define the probability of a set of messages as 
    \[\prob[\sS] := \prod_{(a,b,l) \in \sS} \prob[Z_{ab}^l = 1].\]
\end{definition}

To simplify notation, we will prove all statements in \cref{subsec:WL,subsec:infinite,subsec:oversquashing,subsec:refinement} for $\modelAS$. The proofs for general model architectures $\modelA$ follow completely analogously by performing \ourWL{} on $\graphT = t(\graph)$ instead of $\graph$, i.e., by replacing $\graph$ with $t(\graph)$ and $\modelAS$ with $\modelA$.

\subsection{Function Refinement}
\begin{figure}
    \centering
    \includegraphics[width=0.4\linewidth]{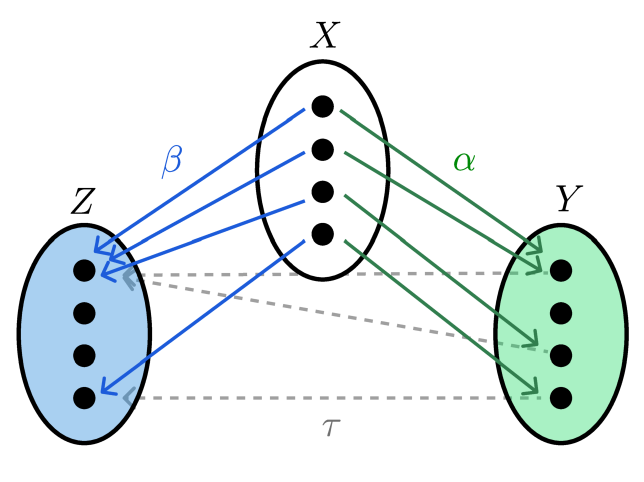}
    \caption{Example for \cref{def:fine_grained} showing sets $X,Y,Z$ and functions $\alpha, \beta$. We have $\alpha \vDash \beta$ as $\alpha(x) \vDash \beta(x)$ for all $x \in X$. By \cref{lem:refinement}, there exists a function $\tau$ with $\beta = \tau \circ \alpha$. Contrarily, $\beta \nvDash \alpha$ as there exist $x_1, x_2 \in X$ with $\beta(x_1) = \beta(x_2)$ but $\alpha(x_1) \neq \alpha(x_2)$. Intuitively, this means that function $\alpha$ is more fine-grained than $\beta$.}
    \label{fig:finegrained_example}
\end{figure}
\begin{figure}
    \centering
    \includegraphics[width=0.4\linewidth]{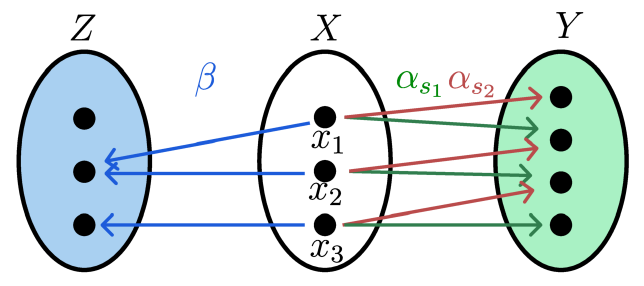}
    \caption{Example for \cref{def:vDashProb} showing sets $X,Y,Z$ and a probabilistic function $\alpha$ and a deterministic function $\beta$. The probabilistic function $\alpha$ can be represented by two deterministic functions $\alpha_{s_1}, \alpha_{s_2}$ where the seed is drawn uniformly at random. We then have $\mathbb{P}[\alpha(x_1) \vDash \beta(x_1)] = 1$. Intuitively, we can deduce $\beta(x_1)$ for every probabilistic outcome of $\alpha(x_1)$. Whereas for $x_2,x_3$, we have  $\mathbb{P}[\alpha(x_2) \vDash \beta(x_2)] = \mathbb{P}[\alpha(x_3) \vDash \beta(x_3)] = 0.5$. }
    \label{fig:finegrained_example_prob}
\end{figure}
First, we will now prove the following useful lemma (see also \cref{fig:finegrained_example}).
\begin{lemma}[Refinement]
    \label{lem:refinement}
    If, and only if, $\alpha \vDash \beta$, there exists a function $\tau$ such that $\beta = \tau \circ \alpha$.
\end{lemma}
\begin{proof}
    First, we will show that $\alpha \vDash \beta$ implies that there exists a function $\tau$ with $\beta = \tau \circ \alpha$. Define $\tau: \alpha(X) \to Z$ as
    \begin{equation*}
        \tau(y) := \beta(x) \:\: \text{for $x \in \alpha^{-1}(y)$}
    \end{equation*}
    This function is well-defined because by definition of $\vDash$, we have for all $x_1, x_2 \in \alpha^{-1}(y)$: $\beta(x_1) = \beta(x_2)$.
    And it follows immediately that $\beta = \tau \circ \alpha$.

    Next, we will show that $\alpha \nvDash \beta$ implies that there does not exist such a function $\tau$ with $\beta = \tau \circ \alpha$. Assume for the sake of contradiction that $\tau$ exists. From $\alpha \nvDash \beta$, it follows that there exist $x_1, x_2$ with $\alpha(x_1) = \alpha(x_2)$ but $\beta(x_1) \neq \beta(x_2)$. Then 
    \[ \tau(\alpha(x_1)) = \tau(\alpha(x_2)) = \beta(x_2) \neq \beta(x_1) = \tau(\alpha(x_1)).\]
    Hence, $\tau(\alpha(x_1)) \neq  \tau(\alpha(x_1))$ which is a contradiction.
\end{proof}

\subsection{Connection of standard MPNNs to WL}
\label{subsec:WL}
Next, we will show that there exists a model in our standard MPNN framework that is at least as expressive as the WL test. In fact, there exists a model performing exactly the WL test:
\begin{lemma}
\label{lem:MWL}
There exists a model $M \in \modelAS$ that performs exactly the WL test, i.e., for all $\graph, v \in \graphsV$ $M_v(\graph) = \WL_v(\graph)$.
\end{lemma}
\begin{proof}
    We show this inductively over the number of layers $L$. For $L=0$ this follows immediately from $\WL^0_v (\graph)= \features_v$ and $M^0_v(\graph) = \features_v$ for any 0-layer model $\model^0 \in \modelA$.
    For $L = l >0$, assume that there exists a model $M^{l-1}$ with $M^{l-1}_v = \WL_v^{l-1}$. Then consider a model $M^l$ that executes $M^{l-1}$ in the first $l-1$ layers:
    \begin{align*}
      m_{w \to v}^l := \begin{cases} \msg_0^l(\WL_w^{l-1}) & \text{if}\:\: w=v \\ \msg_1^l(\WL_w^{l-1}) & \text{else} \end{cases} \quad \text{and}\quad h_v^l := \upd^l (\AGG \left ( \ldbrace  m_{w \to v}^l \mid w \in \neighbor_\graph(v) \cup \{v\} \rdbrace \right ) )
 \end{align*}
 We now need to show that there exist functions $\msg_0^l, \msg_1^l, \AGG$ and $\upd$ such that for all $\graph,v \in \graphs$
\[h_v^l = \WL_v^l = \hash\left(\WL^{l-1}_{v}, \ldbrace \WL^{l-1}_u \mid u \in \neighbor_\graph(v) \rdbrace \right).\] 
\citet{deep_sets} show that one can represent any (permutation-invariant) function $\alpha(X)$ operating on a set $X$ as $\beta\left(\sum_{x \in X} \phi(x)\right)$. As we can differentiate between $\msg_1$, and $\msg_0$ it is easy to see that there also exists functions $\msg_0^l, \msg_1^l$ and $\upd$ such that with sum-aggregation, we have: 
\begin{align*}
\WL_v^l(\graph) &= \hash\left(\WL^{l-1}_{v}, \ldbrace \WL^{l-1}_u \mid u \in \neighbor_\graph(v) \rdbrace \right) \\
&= \upd^l \left(\sum_{w \in \neighbor_\graph(v) \cup \{v\}} \msg^l_{\mathbf{1}[w=v]}(\WL_w^{l-1}) \right ) \\
&= \upd^l \left(\sum_{w \in \neighbor_\graph(v) \cup \{v\}} m_{w \to v}^l \right ) \\
&= h_v^l \\
&= M^l_v(\graph)\qedhere
\end{align*}
\end{proof}

\subsection{When \ours{} becomes infinite: Connections to iso expressivity theory}
\label{subsec:infinite}
We will now prove important theoretical properties of \ours{}. For this, we will first prove the following useful lemma showing that $\ours$ is infinite precisely for the same tasks and graphs where $\WLComp$ is infinite:
\begin{lemma}
We have:
\[\mathbb{P}[\ourWL_v(\graph) \vDash_\graphsV f_v(\graph)] > 0 \]
if and only if
\[\WL_v(\graph) \vDash_\graphsV f_v(\graph).\]
\label{lem:lossyWL0}
\end{lemma}
\begin{proof}
    We will say a message $m_{a\to b}^l$ is visible to $\ourWL_v$ if there exists a $L-l$ length walk from $b$ to $v$.

    We will first assume that $\WL_v(\graph) \vDash_\graphsV f_v(\graph)$. The proof idea for this direction is that with positive probability $\ourWL$ loses no to $v$ visible message and performs exactly the normal WL test. This coloring will always be different from $\ourWL$ colors with visible lost messages and can, hence, by assumption, be used to deduce $f_v$. 
    
    Formally, notice that with probability $>0$ $\ourWL_v^L$ performs exactly the normal Weisfeiler-Lehman test at node $v$, i.e., no to $\ourWL_v$ visible messages are lost ($Z_{uv}^l = 1$ for all visible messages). Now consider such a seed $s_1 \in S$ for which no visible messages are lost. Then for all seeds $s_2 \in S$ and $\graph_2, w \in \graphsV$ for which messages visible to $\ourWL_w(\graph_2)$ are lost, we have \[\ourWL_v(\graph, s_1) \neq \ourWL_w(\graph_2, s_2)\]
    because of the injectiveness of the $\hash$ function, and only $\ourWL_w(t(\graph), s_2)$ has visible lost messages $m = 0$.
    Hence, we have for any $s_2 \in S$ and $\graph_2, w \in \graphsV$ 
        \[\ourWL_v(\graph, s_1) = \ourWL_w(\graph_2, s_2) \implies \WL_v(\graph) = \WL_w(\graph_2)\]
    as $\ourWL_v(\graph, s_1) = \ourWL_w(\graph_2, s_2)$ implies that $s_2$ loses no to $w$ visible messages, i.e., performs exactly the WL test at $w$ as well.
    Therefore,
    \[\ourWL_v(\graph, s_1) \vDash_{\graphsV \times S} \WL_v(\graph).\]
    And from the assumption and the transitivity of $\vDash$, it follows that
    \[\ourWL_v(\graph, s_1) \vDash \WL_v(\graph) \vDash f_v(\graph),\]
    yielding
    \[\mathbb{P}[\ourWL_v(\graph) \vDash_\graphsV f_v(\graph)] > 0 \]

    Now we show that from $\WL_v(\graph) \nvDash_\graphsV f_v(\graph)$, it follows that $\mathbb{P}[\ourWL_v(\graph) \vDash_\graphsV f_v(\graph)] = 0$, i.e., losing messages cannot make the probabilistic WL test more expressive than the deterministic WL test. $\WL_v(\graph) \nvDash f_v(\graph)$ implies there is $\graph', w \in \graphsV$ with 
    $\WL_v(\graph) = \WL_w(\graph')$ but $f_v(\graph) \neq f_w(\graph')$. Then, for every seed $s_1 \in S$, there exists a seed $s_2 \in S$ such that 
    \[\ourWL_v(\graph, s_1) = \ourWL_w(\graph', s_2).\]
    (the same messages to WL equivalent nodes in $\graph$ and $\graph'$ are lost).
    Hence by \cref{def:vDashProb},
    \[\mathbb{P}(\ourWL_v(\graph) \vDash_\graphsV f_v(\graph)) = 0. \qedhere \]
\end{proof}

Using this, we can prove when \ours{} becomes infinite:

\Infeasibility*

\begin{proof}
Assume first that there exists $\graph, v \in \graphsV$ with $\ours_\modelAS(f, \graph) = \infty$. By definition, this implies that 
\[\prob[\ourWL_v(\graph) \vDash_\graphsV f_v(\graph)] = 0.\]
From \cref{lem:lossyWL0} it follows that
\[\WL_v(\graph) \nvDash_\graphsV f_v(\graph).\]
By definition of $\vDash$ this implies that there exists $\graph',w \in \graphsV$ with $\WL_v(\graph) = \WL_w(\graph')$ but $f_v(\graph) \neq f_w(\graph')$. Now it follows from the fact that the WL test upper-bounds the expressivity of standard MPNNs $\modelAS$ \cite{morris} that also all models $M \in \modelA$ cannot differentiate $\graph,v$ from $\graph',w$, i.e., $M_v(\graph) = M_w(\graph')$.

Now, assume that $\ours_\modelAS(f, \graph) \neq \infty$. By definition, this implies that 
\[\prob[\ourWL_v(\graph) \vDash_\graphsV f_v(\graph)] > 0.\]
From \cref{lem:lossyWL0} it follows that
\[\WL_v(\graph) \vDash_\graphsV f_v(\graph).\]
By definition of $\vDash$ this implies that there \emph{do not} exist $\graph',w \in \graphsV$ with $\WL_v(\graph) = \WL_w(\graph')$ but $f_v(\graph) \neq f_w(\graph')$.
Now note that by \cref{lem:MWL} there exists a model $M \in \modelAS$ performing exactly the WL test. Hence, there exists $\model \in \modelAS$ such that there does not exist $\graph',w \in \graphsV$ with $\model_v(\graph) = \model_w(\graph')$ but $f_v(\graph) \neq f_w(\graph')$.
\end{proof}

With this, we can also globally characterize which functions an MPNN can express:

\recover*
\begin{proof}
    If there exists $\model \in \modelA$ such that $\model_v(\graph) = f_v(\graph)$ for all $\graph,v \in \graphsV$, it follows directly from \cref{theo:infeasibility} that there does not exist $\graph,v \in \graphsV$ with $\ours_\modelAS(f_v, \graph) = \infty$.

    If there does not exist $\graph,v \in \graphsV$ with $\ours_\modelAS(f_v, \graph) = \infty$, this implies that for all $\graph,v \in \graphsV$
    \[\prob[\ourWL_v(\graph) \vDash_\graphsV f_v(\graph)] > 0.\]
    Hence, by \cref{lem:lossyWL0} for all $\graph,v \in \graphsV$
    \[\WL_v(\graph) \vDash_\graphsV f_v(\graph).\]
    So
    \[\WL \vDash_\graphsV f.\]
    By \cref{lem:MWL} there exists $M \in \modelAS$ with $M \vDash_\graphsV f$. And by \cref{lem:refinement} a function $\tau$ such that for all $\graph,v \in \graphsV$
    \[\tau \circ M_v(\graph) = f_v(\graph).\]
    But then we can squash the function $\tau$ also in the last update layer of $\model$, i.e., $\upd_{\text{new}}^l = \tau \circ \upd^l$. With this, we have found a model $\model^*$ such that
    \[M^*_v(\graph) = f_v(\graph)\]
    for all $\graph,v \in \graphsV$.
\end{proof}

As a special case, we can recover the iso expressivity statements defined in \cref{def:iso_expressivity}. For this, we use the graph-level formulation of \ours{}. Note that all previous theorems generalize completely analogously also to this graph-level version.
\begin{lemma}[\ours{} preserves impossibility statement of Iso Expressivity]
\label{lem:iso_subsumes}
Let $\alpha$ be a graph isomorphism test. Then $\modelA$ is  at least as expressive as $\alpha$ if and only if there does \emph{not exist} $\graph \in \graphs^*$ with $\ours{}_\modelA(\alpha, G) = \infty$.
\end{lemma}
\begin{proof}
    If $\ours{}_\modelA(\alpha, \graph) = \infty$ for some $\graph \in \graphs^*$, then by \cref{theo:infeasibility}, there exists  $\graph' \in \graphs^*$ such that $\model(\graph) = \model(\graph')$ for all $\model \in \modelA$ but $\alpha(\graph) \neq \alpha(\graph')$. From this follows directly that $\modelA$ is not at least as expressive as $\alpha$.

    Now assume that there does not exist $\graph \in \graphs^*$ with $\ours{}_\modelA(\alpha, \graph) = \infty$. Then by \cref{lem:recover} there exists $M \in \modelA$ with $M(\graph) = \alpha(\graph)$ for all $\graph \in \graphs^*$. Therefore, $\modelA$ is at least as expressive as $\alpha$.
\end{proof}

\subsection{Bounding \ours{}: Function Refinement \& Compositionality}
\label{subsec:refinement}
Next, we will prove bounds on \ours{} relating to how fine-grained a task is and how it can be decomposed into individual subtasks.
First, we prove that a more fine-grained function cannot have lower complexity than the coarser function it refines:
\refinement*
\begin{proof}
    The intuitive idea for this proof is that if the set of successful messages in \ourWL{} suffices to deduce $f$, it is also always possible to deduce the more coarse-grained $g$.
    
    First, note that $\vDash$ is transitive. Then, it follows directly that for every seed $s \in S$ for which $\ourWL_v(G, s) \vDash_{\graphsV \times S} f_v$ also $\ourWL_v(G, s) \vDash_{\graphsV \times S} g_v$.  Therefore
    \[\prob[\ourWL_v(\graph) \vDash_\graphsV f_v(\graph)] \le \prob[\ourWL_v(\graph) \vDash_\graphsV g_v(\graph)] .\]
    And hence,
    \[\ours_\modelA(f_v, \graph) \ge \ours_\modelA(g_v, \graph). \qedhere\]
\end{proof}

Now, we will prove the compositional property of \ours{}.
\triangle*
\begin{proof}
Let $L_f$ be a sufficient set of messages for $f_v$ on graph $G$. Let $\sL_f$ denote the set of all such sufficient sets $L_f$. Notice that $\sL_f$ is upward-closed, i.e., any superset of a $L_f \in \sL_f$ is also sufficient and therefore in $\sL_f$. This intuitively means that more successful messages cannot hurt. Define analogously $\sL_g$ for the task $g$.

For a set of messages $S$ define $\alpha_f(S) = \mathbf{1}[S \in \sL_f]$. Then, $\alpha_f$ is an increasing function over sets, i.e., $S \subseteq S' \implies \alpha_f(S) \le \alpha_f(S')$, because $\sL_f$ is upward-closed. From the Fortuin–Kasteleyn–Ginibre (FKG) inequality, we then have (where $S$ follows the distributions of the sets of successful messages in \ourWL{}):
\[ \mathbb{E}[\alpha_f(S)\alpha_g(S)] \ge \mathbb{E}[\alpha_f(S)] \cdot \mathbb{E}[\alpha_g(S)] \]
implying
\begin{align*}
&\mathbb{P}[\ourWL_v(\graph) \vDash f_v(\graph) \land \ourWL_v(\graph) \vDash g_v(\graph)] \\
&\quad\ge \mathbb{P}[\ourWL_v(\graph) \vDash f_v(\graph)] \cdot \mathbb{P}[\ourWL_v(\graph) \vDash g_v(\graph)] 
\end{align*}
and thereby
\[\ours_\modelA(f_v\Vert g_v, \graph) \le \ours_\modelA(f_v, \graph) + \ours_\modelA(g_v, \graph). \qedhere\]
\end{proof}

\subsection{Connection of \ours{} to Over-squashing and Under-reaching}
\label{subsec:oversquashing}
To formalize \cref{lem:lower_bound}, we first need to define when a task $f_v$ requires information from another node $u$.

\begin{definition}
\label{def:require}
    We say a task $f_v$ over graphs $\graphs$ requires (node-feature) information from a node $u$ if there exists $\graph^1, \graph^2 \in \graphs$ that are identical except for the node features $\features^1_u \neq \features_u^2$ and 
    \[f_v(\graph^1) \neq f_v(\graph^2).\]
\end{definition}

So a task requires information from a node $u$, if without this information it would be impossible to compute $f_v$. 

With this in place, we can give a formal version of \cref{lem:lower_bound}:
\addtocounter{lemma}{1}
\begin{lemma}
\label{lem:lower_bound_formal}
    Consider a task $f_v$ and a graph $\graph \in \graphs$ where $f_v$ requires node-feature information from $u$. Then
    \begin{equation*}
    \ours_\modelA(f_v, \graph) \ge  - \log \left ( (\normInf^L)_{vu} \right ).
    \end{equation*}
\end{lemma}
\begin{proof}
    Intuitively, we will show that in order to deduce $f_v$ from $\ourWL_v$ there needs to exist an L-length walk from $u$ to $v$ where all messages were successful (as $f_v$ requires information from $u$). The probability of this can be in turn upperbounded by the random walk probability.
    
    Formally, we need to show that  
    \[\mathbb{P}[\ourWL_v(\graph) \vDash f_v(\graph)] \le (\normInf^L)_{vu}.\]
    For this we will upper bound $\mathbb{P}[\ourWL_v(\graph) \vDash f_v(\graph)]$ by the $L$ step random walk probability from $v$ to $u$.

    For this, let $W_{ab}$ be the event that there exists $x_1, \dots x_{L-1} \in \vertices$ with 
    \[Z_{ax_1}^1 = 1 \land Z_{x_1 x_2}^2 = 1 \land Z^3_{x_2 x_3} = 1 \land \dots \land Z^{L-1}_{x_{L-2} x_{L-1}} = 1 \land Z^L_{x_{L-1} b} = 1.\]
    We then say $Z$ contains a ($L$-length) walk from $a$ to $b$. Additionally, let $W_{ab}^i$ be the event that $Z$ contains a specific walk from $a$ to $b$ with fully specified intermediate nodes $x_j$. Then the probability of $W_{ab}^i$ is (using the independence of the variables $Z$):
    \begin{align*}
        \prob[W_{ab}^i] &= \prob[Z^1_{ax_1} =1]\cdot \prob[Z_{x_1x_2}^2 = 1]  \cdots  \prob[Z_{x_{L-1}b}^L = 1] \\
        &= \normInf_{x_1a} \cdot \normInf_{x_2x_1} \cdots \normInf_{bx_{L-1}}
    \end{align*}
    This is precisely the probability for a random walk from $b$ to $a$ over all intermediate nodes $x_i$  with edge probabilities $\normInf$.
    If we now consider all $L$-length walks $W_{ab}^1, \dots, W_{ab}^k$ from $a$ to $b$, we get:
    \begin{align*}
        \prob[W_{ab}] &\le \prob[W^1_{ab}] + \cdots + \prob[W_{ab}^k] \\
        &= \prob[\text{$L$-step random walk  from $b$ to $a$ with edge probabilities $\normInf$}] \\
        &= (\normInf^L)_{ba}       
    \end{align*}

    Now note that if \textbf{not} $W_{uv}$, i.e., $Z$ contains no walk from $u$ to $v$, it is by \cref{def:require} not possible to deduce $f_v(\graph)$ from $\ourWL_v(\graph)$ as $v$ will not receive any information from $u$.

    Hence,
    \begin{align*}
    \prob[\ourWL(\graph)_v \vDash f_v(\graph)] &\le \prob[W_{uv}] \\
    &\le (\normInf^L)_{vu} \qedhere.  
    \end{align*}
\end{proof}

With this we can easily prove \cref{cor:under_reaching}:
\begin{proof}
    If $d_\graph(u,v) > L$ there exists no $L$ step walk from $v$ to $u$ on $\graph$. Therefore $(\normInf^L)_{vu} = 0$ and by definition of \ours{}:
    \[\ours_\modelA(f_v, \graph) = \infty.\]
\end{proof}

\subsection{Complexity bounds in \cref{sec:evaluation}}

First, we will prove two useful lemmas to upper and lower bound the \ours{} complexities:

\begin{lemma}
\label{lem:sufficient}
Let $\sS$ be a sufficient set of messages for a function $f_v$ on a graph $\graph$. Then
\[\ours_\modelA(f_v, \graph) \le - \log(\prob[\sS]). \]
\end{lemma}
\begin{proof}
By \cref{def:sufficient} we have:
\begin{align*}
    \prob[\ourWL_v(t(\graph)) \vDash f_v(\graph)] &\ge \prob[\bigwedge_{(a,b,l) \in \sS} Z_{ab}^l = 1] \\
    &= \prob[\sS] \qedhere
\end{align*}
\end{proof}

And a corresponding lower bound:
\begin{lemma}
\label{lem:necessary}
Let $\sS$ be a necessary set of messages for a function $f_v$ on a graph $\graph$. Then
\[\ours(\modelA, f_v, \graph) \ge - \log(\prob[\sS]). \]
\end{lemma}
\begin{proof}
    It follows directly from \cref{def:necessary} that
    \begin{align*}
    \prob[\ourWL_v(t(\graph)) \vDash f_v(\graph)] &\le \prob[\bigwedge_{(a,b,l) \in \sS} Z_{ab}^l = 1] \\
    &= \prob[\sS] \qedhere
\end{align*}
\end{proof}

\paragraph{Retaining information}
We will first prove the complexity bound \cref{lem:retaining} for the task of retaining the initial node feature: $f_v(\graph) = \features_v$.

\retaining*

\begin{proof}
    Note that for $r\ge2$, $\normInf_{ab} < 1$ for all nodes $a,b \in \vertices$ for all model architectures except the MLP. 
    Then, with positive probability, all messages are lost in a layer $l$, i.e., $Z_{a,b}^l = 0$ for all $a,b \in \vertices$. Hence, only with probability $\phi < 1$ any message is successful in layer $l$. 
    A necessary condition for retaining the initial node feature is that in every layer, at least one message is successful.
    Thereby, we have the following lower bound:
    \[\prob[\ourWL_v(t(\graph)) \vDash f(\graph)] \le \phi^L.\]
    It follows immediately, that
    \[\ours(\modelA, f_v, \graph) \ge L \log(1/\phi). \qedhere\]
\end{proof}

\paragraph{Propagating Information}
We will now prove the complexity bound \cref{lem:comp_transfer} for the task $f_v(\graph) = \features_u$ of propagating information from a source node $u$ to a target node $v$ at distance $D = d_\graph(u,v)$. 

\transfer*

\begin{proof}
Trivially, for the \textbf{MLP baseline} and $D>0$, we have:
\[\ours_\text{MLP}(f_v, \graph) = \infty.\]

For \textbf{GCN-VN}, notice that each original node in $\graphT$ has degree $r+2$ ($r$ neighbors in the input graph, the self-loop, and the connection to the virtual node) and the virtual node has degree $1/n$.
It is easy to see that for $D=0$ and $D=1$, the minimal depth required to solve this task is $L=0$ and $L=1$, respectively. Trivially, the $\ours{}$ complexities for these cases are 0 and $\log(r + 2)$. Both are $\le 2 \log(n)$.

For  $D \ge 2$, the minimal depth required to solve this tasks is $L=2$: One sufficient set of messages consists of the message from $u$ to the virtual node (which has probability of success $1/n$) and from the virtual node to $v$ (which has probability of success $1/(r+2)$). By \cref{lem:sufficient}, this implies 
\[\ours_\text{GCN-VN}(f_v, \graph) \le \log(n (r+2)).\]
And hence,
\[\ours_\text{GCN-VN}(f_v, \graph) \le 2\log(n).\]

For all \textbf{other MPNNs}, notice that the probability that $\graph,v \sim \graphsV$ contains a cycle where every node is in the $D$-hop neighborhood around $v$ approaches 0 as $n \to \infty$ for a given $D$. Hence, we can choose $n$ such that this probability is $\le \epsilon$ for any $\epsilon  > 0$.

Then, we can bound the expected complexity by (as \ours{} is always non-negative):
\begin{align*}
 &\mathbb{E}_{\graph,v \sim \graphsV}[\ours_\modelA(f_v,G)]  \\
 &\ge \prob[\text{no cycle around $v$}] \cdot \mathbb{E}_{\graph,v \sim \graphsV}[\ours_\modelA(f_v,G) \mid \text{no cycle around $v$}] \\
 &= (1- \epsilon) \cdot \mathbb{E}_{\graph,v \sim \graphsV}[\ours_\modelA(f_v,G) \mid \text{no cycle around $v$}]
\end{align*}
where "no cycle around $v$" denotes the event that there is no cycle where every node is in the $D$-hop neighborhood around $v$.

Next, we need to bound $\mathbb{E}_{\graph,v \sim \graphsV}[\ours_\modelA(f_v,G) \mid \text{no cycle around $v$}]$. 
For this, we will show that $\ours_\modelA(f_v,G) \ge \log((r+1)^L)$ for any $\graph, v \in \graphsV$ without a cycle around $v$. It is easy to see that for all considered MPNNs without a virtual node, the minimal depth required to solve this task is then $L = D$. Then, a necessary set of messages consists of the messages on the only (as there is no cycle around $v$) $D$-length path from $u$ to $v$. Each message on this path has a success probability $\le 1/(r+1)$. By \cref{lem:necessary}, this implies:
\[\ours_\modelA(f_v,G) \ge \log((r+1)^D)\]
Finally, by choosing $\epsilon$ small enough, we have
\begin{align*}
&\mathbb{E}_{\graph,v \sim \graphsV}[\ours_\modelA(f_v,G)]  \\
&\ge (1- \epsilon) \cdot \mathbb{E}_{\graph,v \sim \graphsV}[\ours_\modelA(f_v,G) \mid \text{no cycle around $v$}] \\
&\ge (1- \epsilon) \cdot \log((r+1)^D) \\
&\ge D \log(r) \qedhere
\end{align*}
\end{proof}

\paragraph{Extracting topological information}
Lastly, we consider the ring transfer task. To simplify notations here, we will consider the complexity of $f_v(\graph) = \features_u$ for a \emph{single node} $u$ on the cycle of size $s$ that $v$ is part of. The tasks involve 1) identifying that $u$ is part of the cycle 2) propagating the node feature information from $u$ to $v$. We will now prove the complexity bounds in \cref{lem:ring_bounds}:

\begin{lemma}
\label{lem:ring_prop_formal}
Assume $L$ is the minimal depth required to solve the task $f_v(\graph) = \features_u$ with architecture $\modelA$  where $u$ is a node on the cycle of size $s$ that $v$ is part of. Further assume that there is only a single cycle in the $\lceil s/2 \rceil$-hop neighborhood of $v$. Then, the expected \ours{} complexities are: 
\[\text{For CIN \& FragNet:}\:\: O(\log(sr)) \quad \quad \text{For GSN:} \:\: \leq \lceil s/2 \rceil \log(r+1) \quad\quad \text{For all others:}\:\: \geq s \log(r)\]
\end{lemma}

\begin{proof}
\textbf{FragNet:} Notice that for FragNet $L=2$ is the minimal depth required to solve this task. With $L = 2$ a sufficient set of messages involves a message from  $u$ to the fragment node that represents this cycle (with message success probability $1/s$) and from the fragment node to $v$ (with message success probability $1/(r+2)$). Using \cref{lem:sufficient}, the bound follows immediately.

\textbf{CIN:} Again, for CIN $L=2$ is the minimal depth required to solve this task.
A sufficient set of messages involves a message from the ring node (which contains as initial encoding all node features from the nodes in the ring) to an edge node. And from the edge node to $v$. Notice that the edge node has degree $s+2$ (an edge to every other edge node in the cycle, an edge to the ring node and two edges to the original nodes) and $v$ has degree $2r+1$ (edges to all $r$ neighboring nodes in $\graph$, edges to corresponding edge nodes, and self-loop). Hence, we have message success probabilities $1/(s+2)$ and $1/(2r+1)$. Again, using \cref{lem:sufficient}, the bound follows immediately.

\textbf{GSN:} Notice that for GSN the cycle information is already contained in the initial node-feature. So the task is simply to propagate the node feature information to $v$ for a distance of at most $\lceil s/2\rceil$. Hence, a sufficient set of messages consists of the messages on a path of length $L \le \lceil s/2 \rceil$ from $u$ to $v$, each with success probability $1/(r+1)$. Again, using \cref{lem:sufficient}, the bound follows immediately.

\textbf{All other MPNNs:} For all other MPNNs, the minimum depth required to solve this task is $L=s$ (for $L < s$, it is impossible to detect the cycle). It is easy to see that in order to identify the cycle at node $v$, there needs to be a set of successful messages going around the full cycle returning to $v$. Or more formally, for a cycle consisting of nodes $v, a_1, a_2, \dots, a_{s-1}, v$, we define two possible message traversal patterns:
\[\text{Cycle}_\circlearrowright := Z_{va_1}^1 = 1 \land Z_{a_1a_2}^2= 1 \land Z_{a_2a_3}^3=1 \land \cdots \land Z_{a_{s-1}v}^s=1\]
and
\[\text{Cycle}_\circlearrowleft := Z_{va_{s-1}}^1 = 1 \land Z_{a_{s-1}a_{s-2}}^2= 1 \land Z_{a_{s-2}a_{s-3}}^3=1 \land \cdots \land Z_{a_{1}v}^s=1.\]
Additionally, node $v$ must retain its unique node feature throughout all $L$ layers to recognize that the returning messages originated from itself.\footnote{A sufficient condition for this task would be even stronger, requiring all nodes in the cycle to retain their features until the  messages through the cycle reach them.}
Hence, a necessary condition for $\ourWL_v(\graph) \vDash f_v(\graph)$ is 
\[Z_{vv}^1 = 1 \land Z_{vv}^2=1 \land \cdots \land Z_{vv}^s=1 \land (\text{Cycle}_\circlearrowright \lor \text{Cycle}_\circlearrowleft).\]
Therefore, we have 
\[\prob[\ourWL_v(\graph) \vDash f_v(\graph)] \le  \prob[Z_{vv}^1 = 1 \land Z_{vv}^2=1 \land \cdots \land Z_{vv}^s=1 \land (\text{Cycle}_\circlearrowright \lor \text{Cycle}_\circlearrowleft)].\]
Notice that $\prob[Z_{ab}^l] = 1/(r+1)$ for any message, as every node has degree $r$ and the additonal self-loop. Then, because of the independence of all $Z_{ab}^l$, we have
\begin{align*}
\prob[\ourWL_v(\graph) \vDash f_v(\graph)] &\le  \prob[Z_{vv}^1 = 1 \land Z_{vv}^2=1 \land \cdots \land Z_{vv}^s=1 \land (\text{Cycle}_\circlearrowright \lor \text{Cycle}_\circlearrowleft)]. \\
&= \prob[Z_{vv}^1 = 1 \land Z_{vv}^2=1 \land \cdots \land Z_{vv}^s=1] \cdot  \prob[\text{Cycle}_\circlearrowright \lor \text{Cycle}_\circlearrowleft] \\
&\le 1/2 \cdot \prob[\text{Cycle}_\circlearrowright \lor \text{Cycle}_\circlearrowleft)] \\
&\le 1/2 \cdot 2 \cdot \prob[\text{Cycle}_\circlearrowright] \\
&= \left(\frac{1}{r+1}\right)^s.
\end{align*}
While this bound is not tight (we've made several relaxations), it is sufficient to establish the $\ge s\log(r)$ \ours{} complexity lower bound.

\textbf{MLP:} The complexity is trivially infinite.
\end{proof}

Lastly, we give a short proof why $\WLComp$ is $0$ for this task. For this, we generally show that for graphs with unique node features any (at least WL-expressive) architecture with at least one layer can distinguish all non-isomorphic graphs, and hence solve any task.

\begin{lemma}
    Let $\graphs$ be a family of graphs with unique node features, i.e., for any $\graph \in \graphs$, no two nodes $v,w \in \vertices$ have the same node features. Then the WL output after one iteration $\ldbrace \WL_v^1 \mid v \in \vertices \rdbrace$ differs for any two non-isomorphic graphs in $\graphs$.
\end{lemma}
\begin{proof}
    Let $\graph_1, \graph_2 \in \graphs$ be two non-isomorphic graphs. Suppose, for contradiction, that the WL output after one iteration is the same for both graphs, i.e., $\ldbrace \WL_v^1 \mid v \in \vertices_1 \rdbrace = \ldbrace \WL_w^1 \mid w \in \vertices_2 \rdbrace$ as multisets.
    
    Since all node features in $\graph_1$ and $\graph_2$ are unique within each graph, the initial WL labels $\WL_v^0$ are distinct for all nodes $v$ in each graph. After one iteration, each $\WL_v^1$ consists of the node's original feature combined with the multiset of its neighbors' features, i.e., $\WL_v^1 = \left(\WL_v^0, \ldbrace \WL_u^0 \mid u \in \neighbor(v) \rdbrace \right)$.
    
    Given that $\ldbrace \WL_v^1 \mid v \in \vertices_1 \rdbrace = \ldbrace \WL_w^1 \mid w \in \vertices_2 \rdbrace$, there must exist a bijection $f: \vertices_1 \rightarrow \vertices_2$ such that $\WL_v^1 = \WL_{f(v)}^1$ for all $v \in \vertices_1$.
    
    This implies that for each $v \in \vertices_1$:
    (1) $\WL_v^0 = \WL_{f(v)}^0$ (the original node features match), and 
    (2) $\ldbrace \WL_u^0 \mid u \in \neighbor(v) \rdbrace = \ldbrace \WL_w^0 \mid w \in \neighbor(f(v)) \rdbrace$ (the multisets of neighbor features match).
    
    From (1), since node features are unique within each graph, $f$ maps each node in $\graph_1$ to a unique node in $\graph_2$ with identical features.
    
    From (2), for each $v \in \vertices_1$, there must exist a bijection $g_v: \neighbor(v) \rightarrow \neighbor(f(v))$ such that $\WL_u^0 = \WL_{g_v(u)}^0$ for all $u \in \neighbor(v)$.
    
    Due to the uniqueness of node features and (1), we must have $g_v(u) = f(u)$ for all $u \in \neighbor(v)$. This implies that for all $v \in \vertices_1$ and $u \in \vertices_1$:
    $u \in \neighbor(v) \iff f(u) \in \neighbor(f(v))$
    Therefore, $f$ is an isomorphism between $\graph_1$ and $\graph_2$, contradicting our assumption that the graphs are non-isomorphic.
    Thus, the WL output after one iteration must differ for any two non-isomorphic graphs in $\graphs$.
\end{proof}

\section{Additional results and Experimental Details}
We present the following additional results:
\begin{itemize}
    \item Extended figures showing \emph{all} considered model architectures across all tasks.
    \item Robustness analysis demonstrating that \ours{} predictions hold across different hyperparameter choices (hidden dimensions) for the information retention task.
    \item Validation on Erdős-Rényi (ER) graphs, confirming that \ours{} accurately matches performance trends also on a different synthetic graph family.
    \item Real-world validation on graphs from ZINC \cite{ZINC} and peptides datasets \cite{lrgb} showing that \ours{} explains performance better than classical expressivity theory for real-world graph structures.
    \item Results showing that insight on synthetic proxy tasks can mirror performance trends on real-world benchmarks where the precise tasks are unknown.
\end{itemize}
Additionally, we carefully describe all experimental details and setups.

\subsection{Implementation Details and Experimental Setup}
\label{app:experimental_details}
We use pytorch \cite{pytorch} and torch-geometric \cite{pytorch_geometric} (released under an MIT license) for the implementation of all models. For optimization, we use the Adam optimizer \cite{adam}. The training, validation and test sets containing random $r$-regular graphs are generated using networkx \cite{networkx}.
For all standard MPNNs we use their pytorch geometric implementation. Note that GSN \cite{gsn} only specifies the additional node features and not the downstream MPNN. We use the most common MPNN, GCN, as downstream MPNN for all experiments for GSN. For CIN \cite{cw} we use our own implementation following exactly the method proposed in their paper. Additionally, for FragNet, we use a custom implementation of their FR-WL model without edge representations using a GCN as base MPNN. For CIN, GSN, and FragNet we use a fragmentation scheme identifying every cycle of size at most 6 (unless otherwise noted).
For all models we use an initial feature embedding layer and a final output MLP. Additionally, we use BatchNorm \cite{batch_norm} for normalization. We found little difference between learning rates in $\{0.001, 0.005, 0.01\}$ for all models and tasks. The shown results are for the learning rate $0.005$.  We train all models for all settings using three different seeds for a maximum of 50 epochs, showing the average results. All experiments are conducted on NVIDIA GeForce GTX 1080 GPUs with 16GB memory allocation per job. Training times vary by model architecture, ranging from 10-30 minutes. Monte Carlo simulations for complexity calculations run on a single Intel Xeon E5-2630 v4 CPU (2.20GHz) and complete in under 10 seconds. The peptides dataset of the lrgb \cite{lrgb} is released under a CC BY-NC 4.0 license. The ZINC dataset \cite{ZINC} is distributed under a custom license (free to use for everyone).

\subsection{Retaining information}
\begin{figure}
    \centering
    \includegraphics[width=1\linewidth]{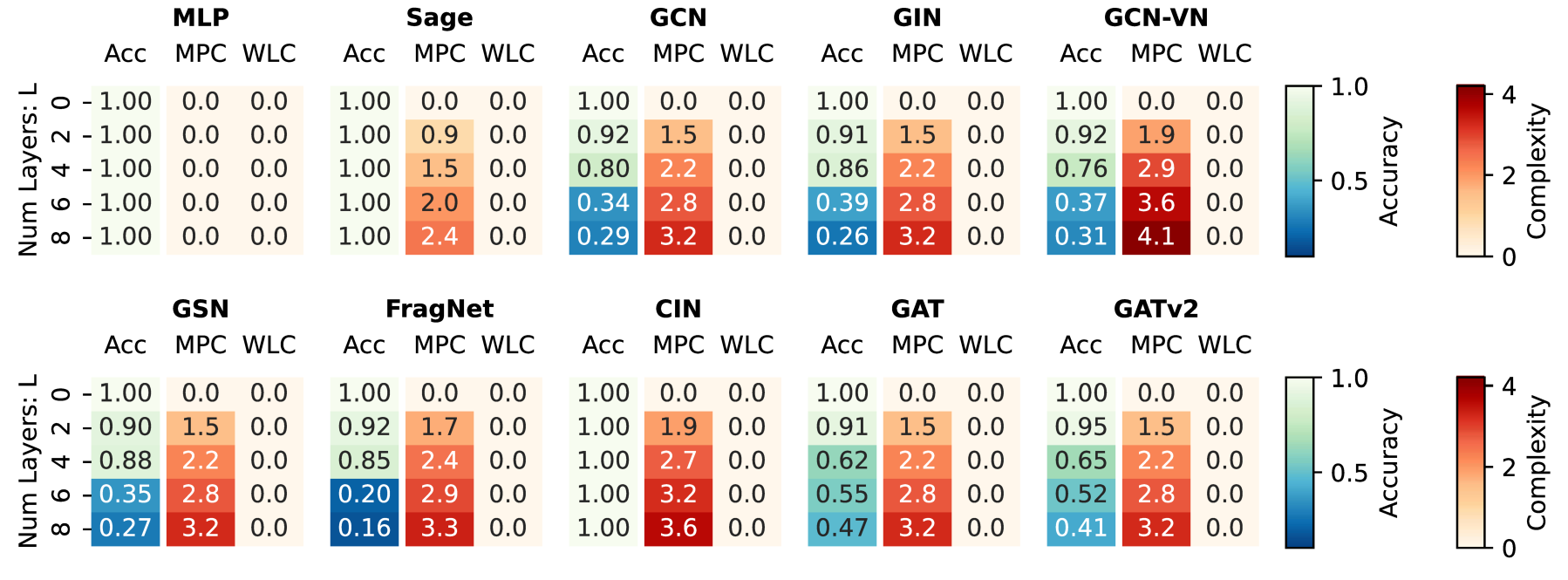}
    \caption{Test accuracy for retaining initial node features compared with complexity measures \ours{} and \WLComp{} for all models. Simulated \ours{} (in contrast to WL-based \WLComp{}) matches trends in empirical accuracy (for all models except for Sage and CIN), capturing increasing difficulty with depth.}
    \label{fig:keep_heatmap_full}
\end{figure}

\begin{figure}
    \centering
    \includegraphics[width=1\linewidth]{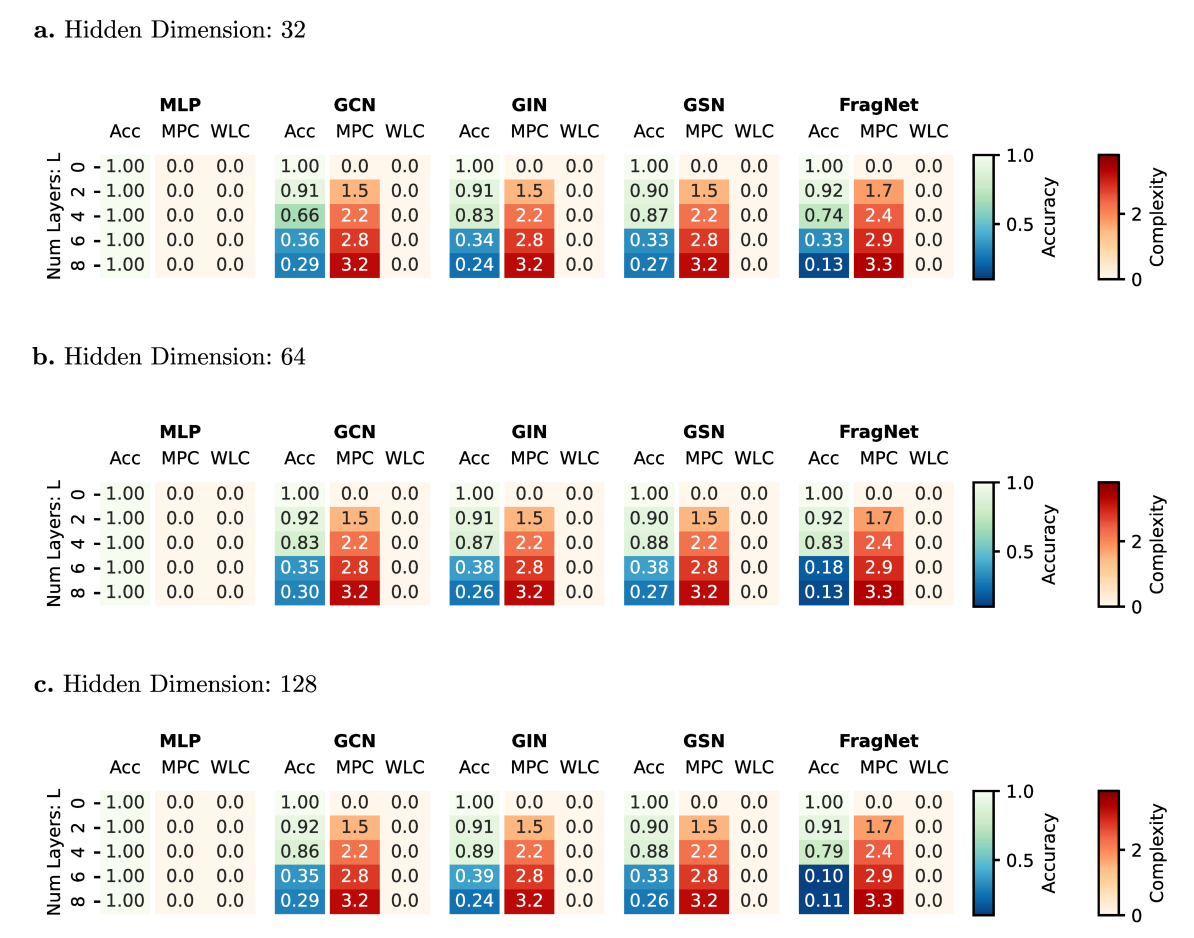}
    \caption{Influence of hidden dimension hyperparameter. Test accuracy for retaining initial node feature task compared with complexity measures \ours{} and \WLComp{} for hidden dimension size \textbf{a.} 32, \textbf{b.} 64 and \textbf{c.} 128 (for a selection of models). Even with larger hidden dimensions, MPNNs face fundamental over-smoothing limitations that \ours{} captures while \WLComp{} does not. \ours{} consistently predicts performance trends across all hidden dimension choices.}
    \label{fig:keep_hyperparameter}
\end{figure}

\begin{figure}
    \centering
    \includegraphics[width=1\linewidth]{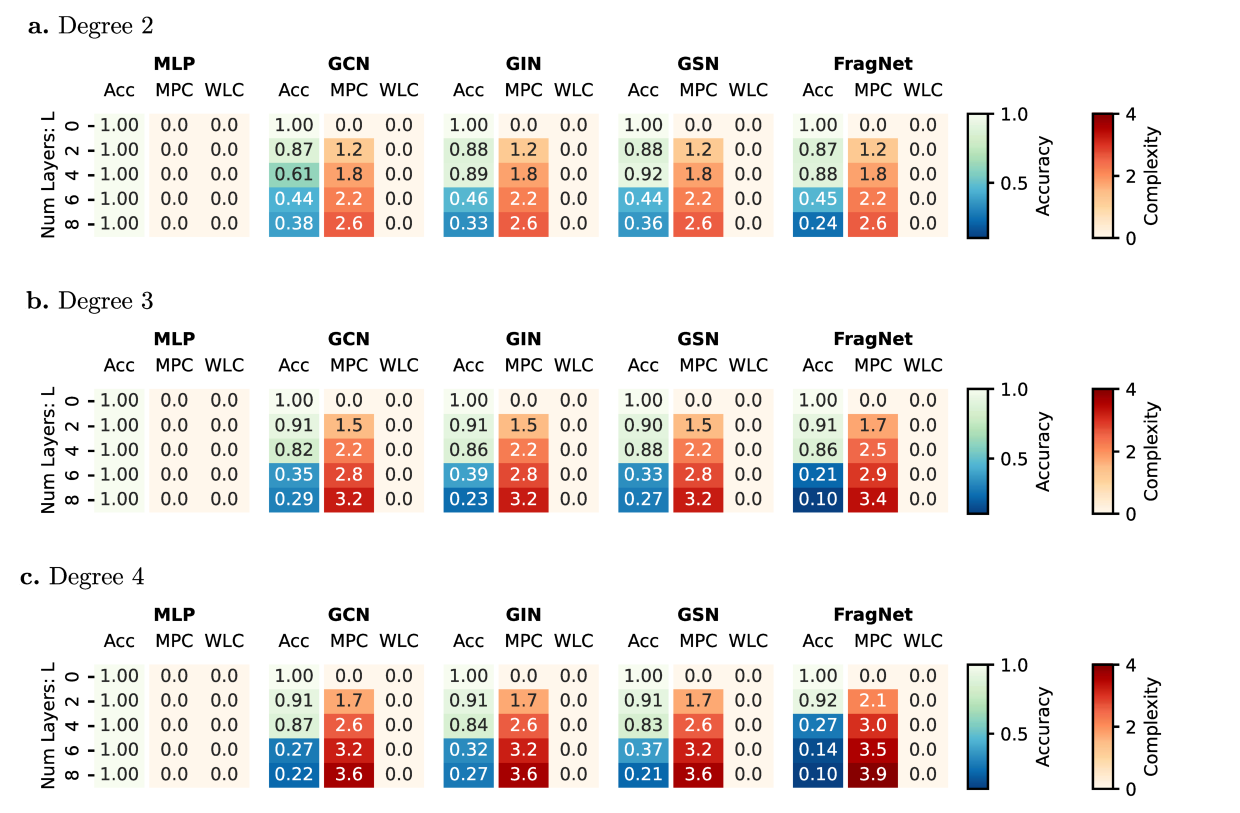}
    \caption{Influence of degree $r$ of the random regular graphs.  Test accuracy for retaining initial node feature task compared with complexity measures \ours{} and \WLComp{} for hidden dimension size \textbf{a.} 32, \textbf{b.} 64 and \textbf{c.} 128 (for a selection of models). Unlike \ours{}, MPC captures the increasing difficulty as graph degree increases, demonstrating the importance of considering the graph topology.}
    \label{fig:keep_dense_sparse}
\end{figure}

For the task $f_v(\graph) = \features_u$, we randomly generate $3$-regular graphs with $n=50$ nodes where each node is randomly assigned to one of ten possible classes. We use a training set of size 2000, a validation set of size 500 and a test set of size 2000. 
We simulate the complexities using Monte Carlo simulation on 100 different graphs from the test set, and for each graph 1000 trials (with the same method explained in \cref{subsec:propagating}).
\cref{fig:keep_heatmap_full} shows the accuracy and simulated \ours{} complexity for all considered model architectures. Additionally, \cref{fig:keep_hyperparameter} shows that the complexity measure is robust to changes in the hidden dimension hyperparameter, i.e., trends in \ours{} complexity align with empirical performance also for different hidden dimensions. Moreover, \cref{fig:keep_dense_sparse} shows how complexity and accuracy changes for different levels of sparsity.

\begin{figure}
    \centering
    \includegraphics[width=1\linewidth]{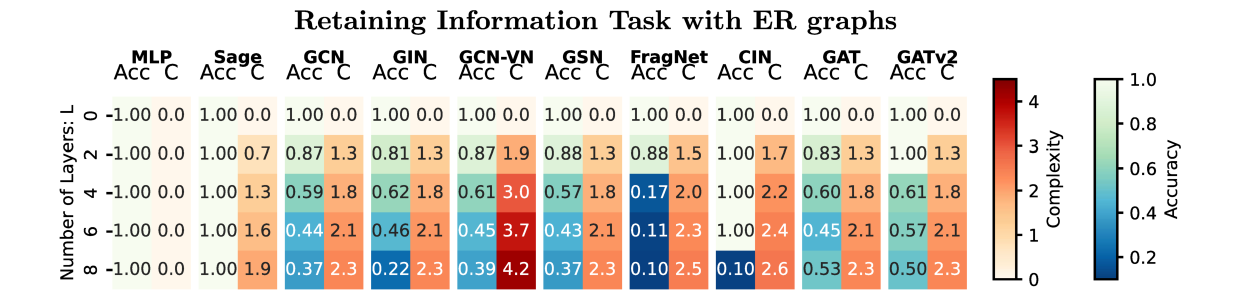}
    \caption{MPC captures practical difficulty across different graph families. Test accuracy and average simulated MPC complexity for retaining initial node features on Erdős–Rényi graphs (50 nodes, connection probability p = 0.06). Models are trained on 2000 graphs. MPC complexity (C) correctly increases with layer depth (\WLComp{} would be 0 everywhere), demonstrating that our complexity measure generalizes beyond random regular graphs to capture practical learning difficulties.}
    \label{fig:keep_er}
\end{figure}

\paragraph{Erdős–Rényi graphs}
We repeat the retaining information experiment with Erdős–Rényi (ER) graphs with $n=50$ nodes and node connection probability $p = 0.06$. \cref{fig:keep_er} shows that generally, trends in \ours{} complexity align with trends in empirical performance (except for Sage and CIN, the two models that directly optimize the weight of the residual channel) as well for ER graphs. 

\clearpage
\subsection{Propagating Information}
\label{subsec:propagating}
\begin{figure}
    \centering
    \includegraphics[width=1\linewidth]{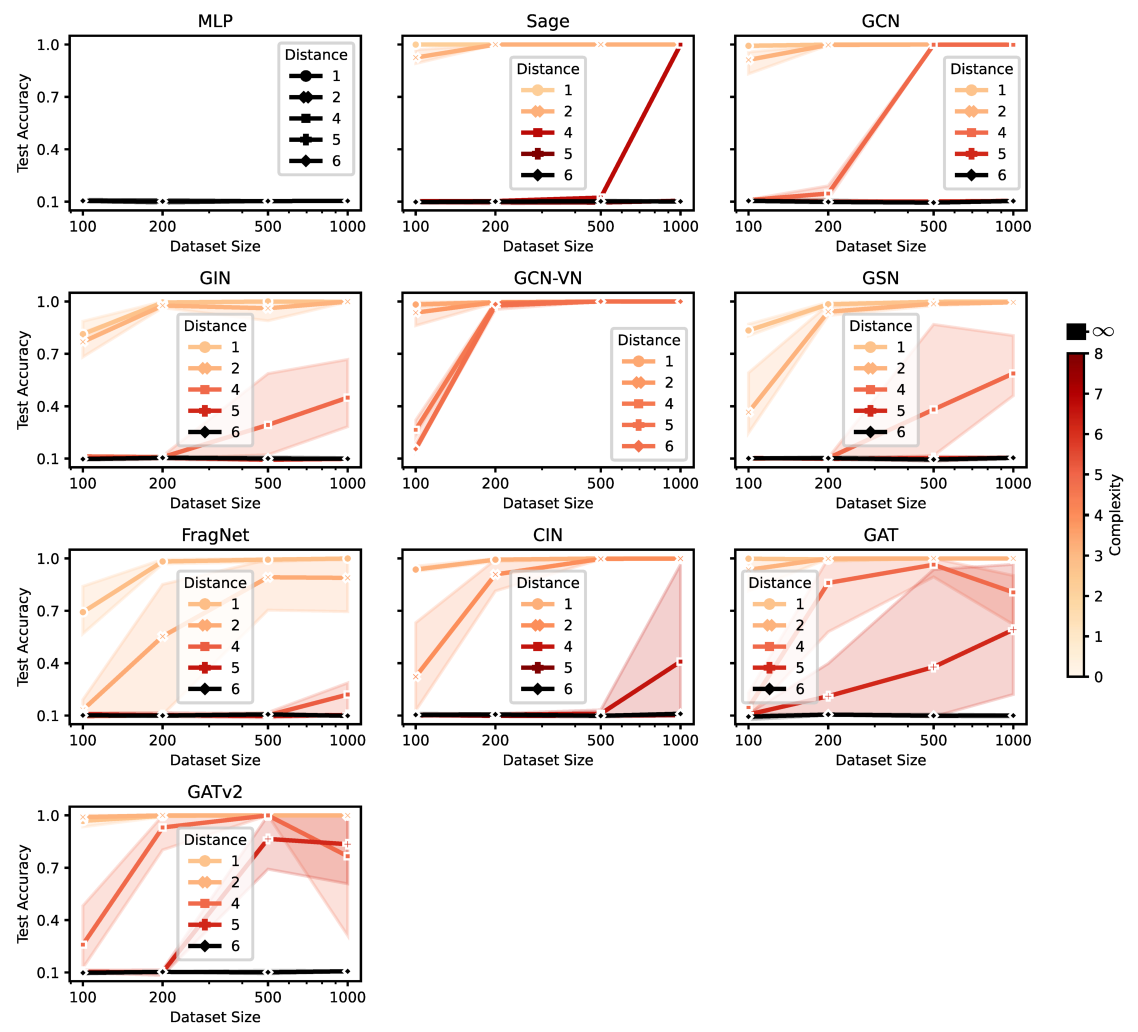}
    \caption{Test accuracy by training data size for the information propagation task $f_v(\graph) = \features_u$ for different distances $D$. Colored by average simulated \ours{} complexity per distance. \ours{} correctly captures the increasing sample complexity with distance and the performance advantage that a virtual node offers for long-range dependencies.}
    \label{fig:transfer_full}
\end{figure}

\begin{figure}
    \centering
    \includegraphics[width=1\linewidth]{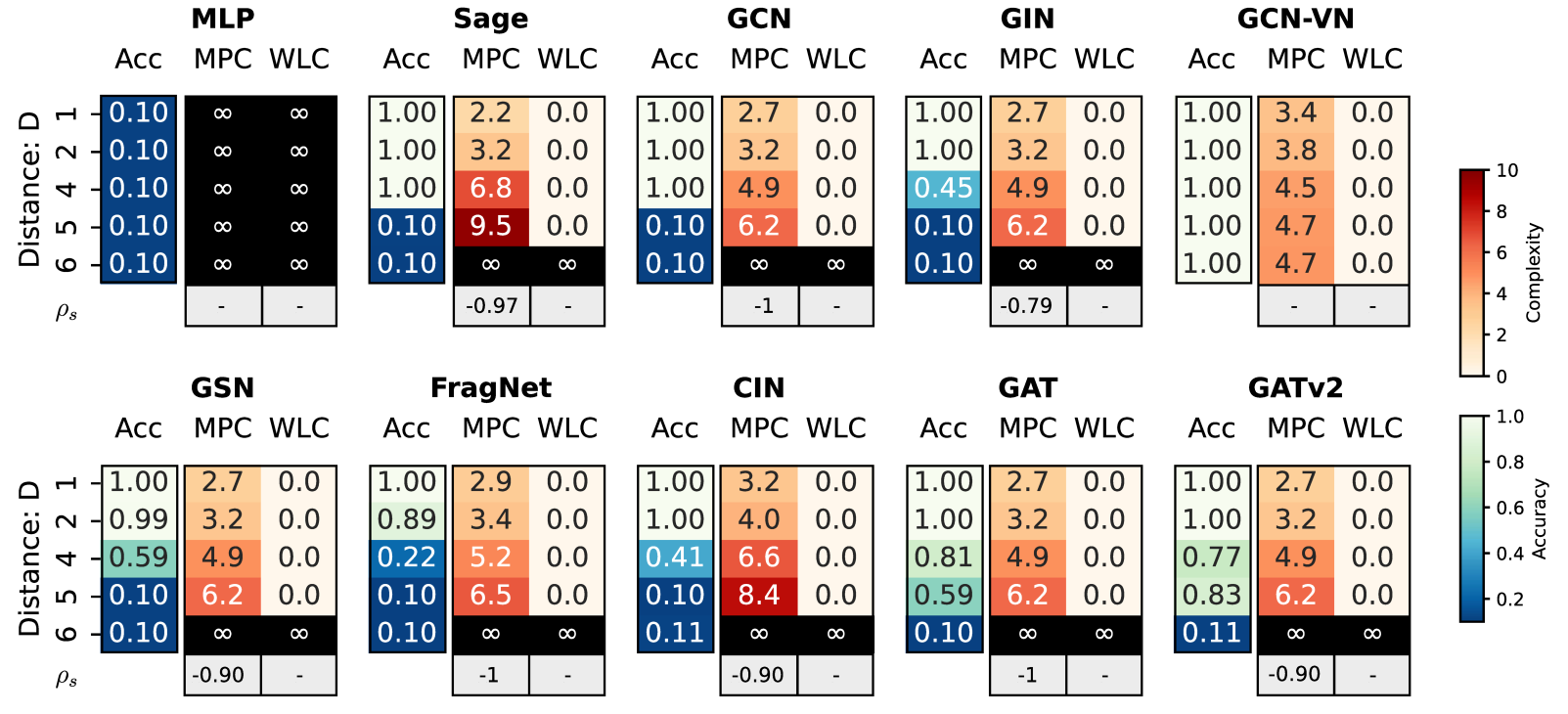}
    \caption{Test accuracy for retaining initial node features compared with complexity measures \ours{} and \WLComp{}  for the information propagation task $f_v(\graph) = \features_u$ for all models (for dataset size 1000). Simulated \ours{} matches trends in empirical accuracy (highly negative Spearman correlated $\rho_S$), capturing increasing difficulty with distance while preserving impossibility statements from $\WLComp{}$.}
    \label{fig:transfer_full_heatmap}
\end{figure}

For the task $f_v(\graph) = \features_u$, we randomly generate 3-regular graphs with $n=50$ nodes. In each graph, we randomly select a target node $v$. For a given distance $D$, we select a node $u$ with distance $D$ from $v$. Then, $v$ gets the unique label $0$ identifying it as target node, $u$ is randomly assigned a target label in $\{1, \dots, 10\}$, and all other nodes get a random label from $\{11, \dots, 20\}$. For each distance $D$, we use training sets of different sizes and a validation set of size 500 and a test set of size 2000. 
We simulate the complexities using Monte Carlo simulation on 100 different graphs from the test set, and for each graph 1000 trials.
\cref{fig:transfer_full} shows the average test accuracy by training data size and \cref{fig:transfer_full_heatmap} shows the test accuracy and the complexity for the dataset size 1000.

\begin{figure}
    \centering
    \includegraphics[width=1\linewidth]{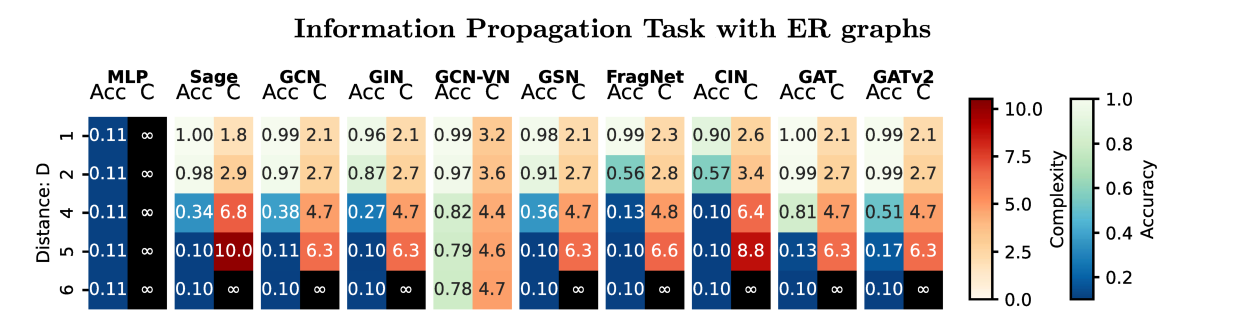}
    \caption{Test accuracy by distance for the information propagation task $f_v(\graph) = \features_u$ (for varying train dataset sizes) and simulated \ours{} complexity (C) for ER graphs with $n=50$ nodes and connection probability $p = 0.04$. MPC again matches empirical performance trends, demonstrating that our complexity measure generalizes beyond random regular graphs to capture practical learning difficulties.}
    \label{fig:transfer_er_heatmap}
\end{figure}

\paragraph{Erdős–Rényi graphs}
Again, we repeat the information propagation experiment with ER graphs with $n=50$ nodes and node connection probability $p = 0.04$. \cref{fig:transfer_er_heatmap} show that trends in \ours{} complexity align with trends in empirical performance as well for ER graphs. 

\stepcounter{figure}

\paragraph{Peptides dataset}
\begin{figure}
    \centering
    \includegraphics[width=1\linewidth]{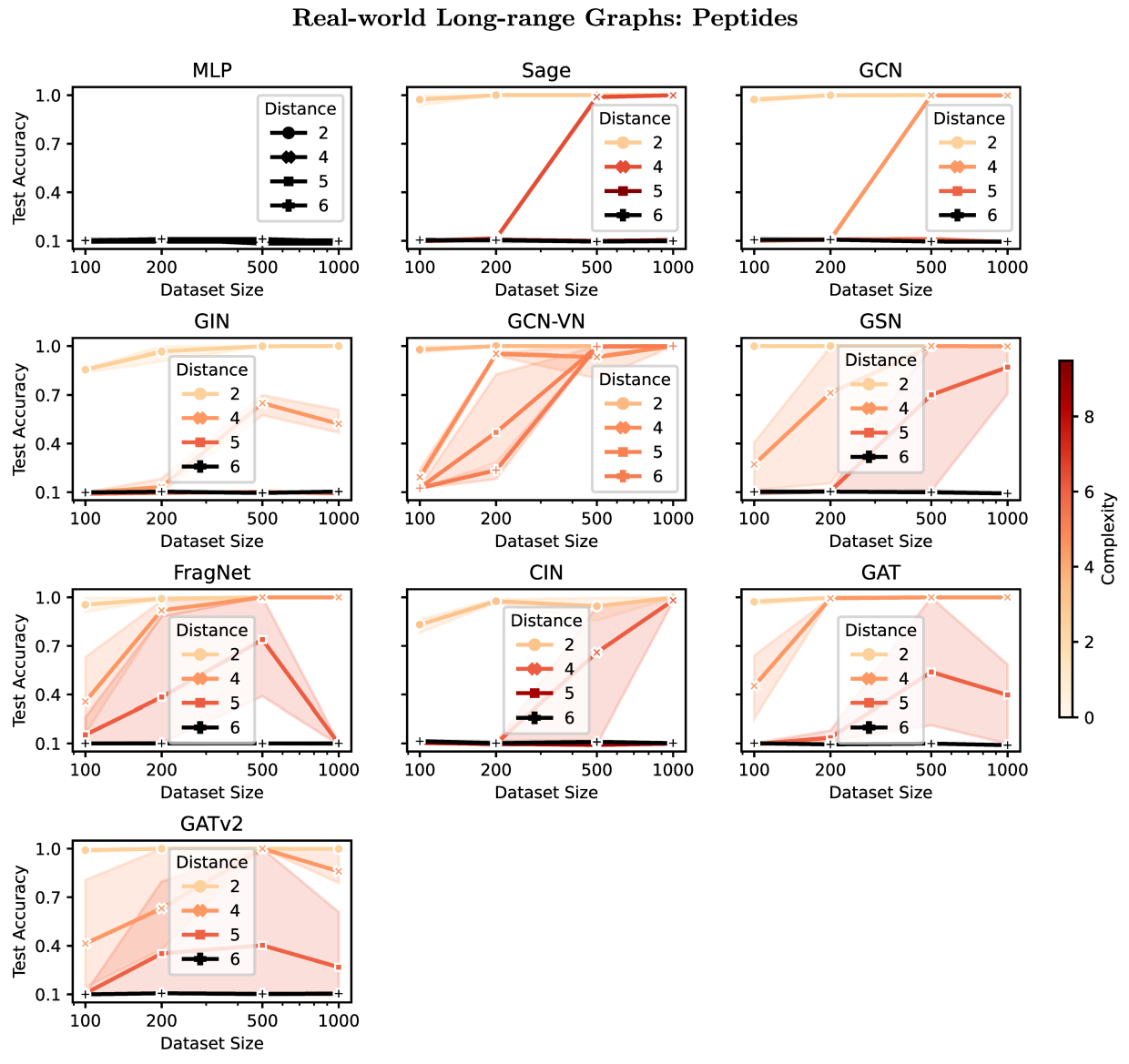}
    \caption{\textbf{\ours{} captures task difficulty on real-world graphs.} Test accuracy by distance for the information propagation task $f_v(\graph) = \features_u$ across varying training dataset sizes, with simulated \ours{} complexity shown for real-world graphs from the peptides-func/peptides-struct dataset \cite{lrgb}. Higher \ours{} complexity corresponds to greater learning difficulty, as evidenced by increased sample complexity requirements, demonstrating that \ours{} effectively captures practical task difficulty beyond synthetic graph families.}
    \label{fig:peptides_line}
\end{figure}
Additionally, we repeat the information propagation task on graphs from the real-world peptides-func/peptides-struct dataset. It comprises larger peptide molecules that require models to consider long-range interactions \cite{lrgb}. Therefore, the information propagation task, testing the models' ability to exchange information over varying distances, is particularly well-suited for this dataset. 

Similar to our approach for random regular and ER graphs, we randomly select one node of each graph in the dataset as target node $v$. For a given distance $D$, we randomly select a node $u$ with distance $D$ from $v$ as the source node. Then, $v$ gets the unique label $0$ identifying it as target node, $u$ is randomly assigned a 
label in $\{1, \dots, 10\}$, and all other nodes get a random label from $\{11, \dots, 20\}$. We then randomly sample different numbers of graphs from the dataset as train graphs. In summary, this allows us to analyze and isolate the information propagation capability of models on real-world graphs requiring long-range interactions. 

\cref{fig:peptides_line} shows again a connection between sample complexity needed to achieve perfect accuracy and MPC complexity, underlining that MPC is also a good predictor of performance for graph topologies occurring in the real world.
\clearpage

\subsection{Extracting topological information}
\begin{figure}[b]
    \centering
    \includegraphics[width=1\linewidth]{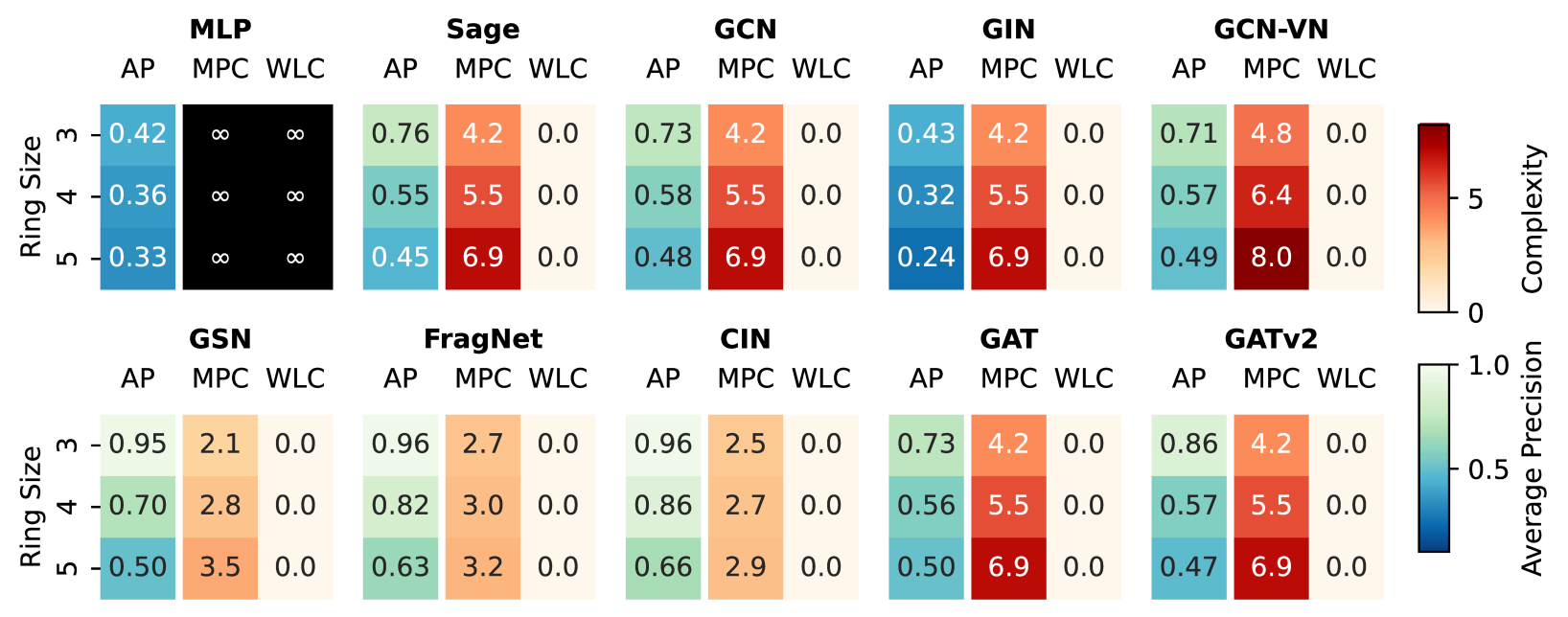}
    \caption{Test average precision for the ring transfer task compared with complexity measures \ours{} and \WLComp{} across all models (dataset size 10000, \ours{} values for all standard MPNNs are lower bounds). Simulated \ours{} aligns with empirical accuracy trends, capturing both the increasing difficulty with ring size and the superior performance of GSN, FragNet, and CIN due to their cycle-oriented inductive biases.}
    \label{fig:ring_heatmap_full}
\end{figure}

\begin{figure}
    \centering
    \includegraphics[width=1\linewidth]{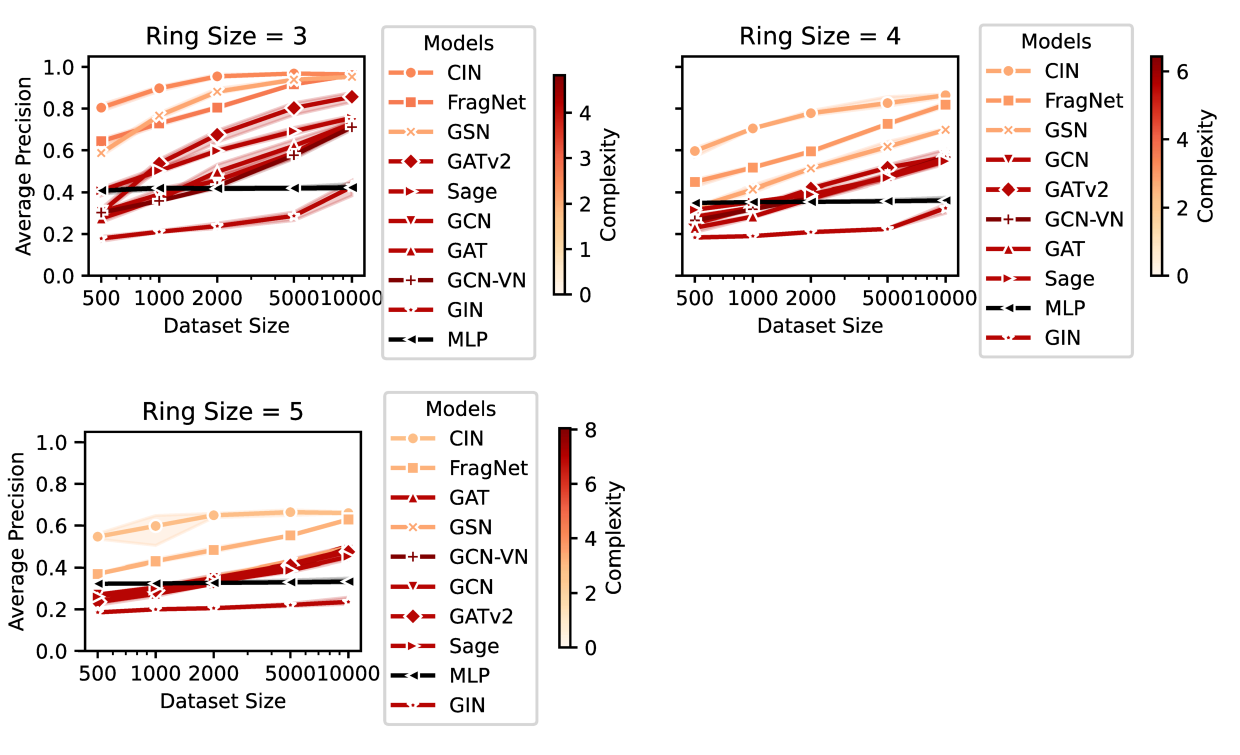}
    \caption{Test average precision and by dataset size for the ring transfer task in relation to dataset size. Models are colored by their simulated \ours{} complexity.}
    \label{fig:ring_line_full}
\end{figure}

For the ring transfer task, we use randomly generated $4$-regular graphs with $n=50$ nodes conditioned on having a cycle of size $s$ at a node $v$ (and no smaller cycle). Each node is randomly assigned a unique label in $\{1, \dots, 50\}$. The multilabel classification task $f_v(\graph)$ is then to classify which labels between 1 and 50 are part of the cycle that contains $v$. For each cycle size, we generate training sets of varying size, a validation set of size 1000 and a test set of size 10000. For GSN, FragNet, and CIN we use a fragmentation scheme identifying cycles of size at most 5. For all models, we use the minimal number of layers with which they can solve all tasks: for FragNet and CIN: 2, for all other models: 5.
For CIN and FragNet, we explicitly compute the complexities. For all other models, we use the bounds provided in \cref{lem:ring_bounds}.
All results show the binary average precision. \cref{fig:ring_heatmap_full} shows the average precision and complexities for this task for the maximal dataset size of 10000. Additionally, \cref{fig:ring_line_full} shows the average precision in relation to dataset size.

\begin{figure}
    \centering
    \includegraphics[width=1\linewidth]{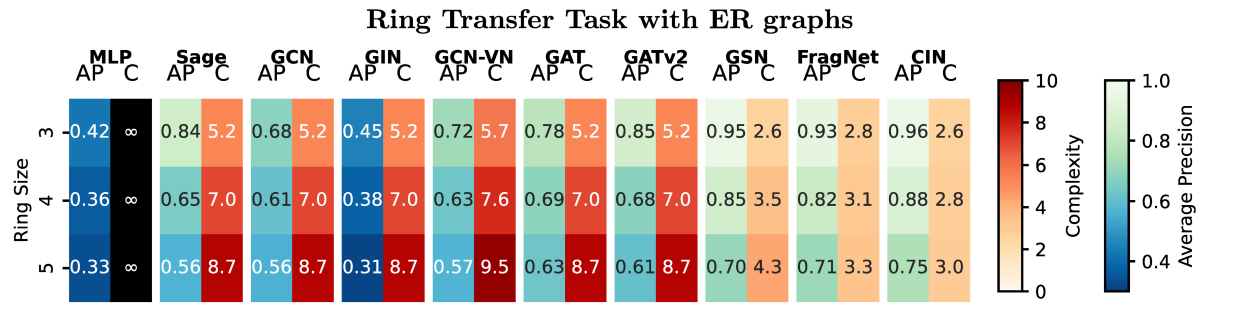}
    \caption{Test average precision and average \ours{} complexity (C) for the ring transfer task with ER graphs for all models for training data size 10000 (complexity values for standard MPNNs are lower bounds), demonstrating again that \ours{} matches performance trends for graphs from a different distribution.}
    \label{fig:ring_er_heatmap}
\end{figure}

\paragraph{Erdős–Rényi graphs}
Again, we repeat the ring transfer task with ER graphs with $n=50$ nodes and edge probability $p=0.04$ conditioned on having a ring of size $s$ at node $v$. The label assignment is done in the same way as for the random regular graphs. \cref{fig:ring_er_heatmap} shows that complexities also align with MPC complexities for ER graphs.

\stepcounter{figure}

\paragraph{ZINC dataset}
\begin{table}[tbp]
\centering
\caption{Comparison of MPC for ring transfer task (for ring size 6) and 
Mean Absolute Error on prediction penalized logP, both on graphs from ZINC-subset. Trends in MPC on the synthetic ring task match trends on the real-world task, where identifying large rings is important.}
\begin{tabular}{lcccccc}
\toprule
& \multicolumn{3}{c}{\textbf{Standard MPNNs}} & \textbf{Substructure} & \multicolumn{2}{c}{\textbf{Substructure}} \\
& \multicolumn{3}{c}{} & \textbf{Encodings} & \multicolumn{2}{c}{\textbf{Graphs}} \\
\cmidrule(lr){2-4} \cmidrule(lr){5-5} \cmidrule(lr){6-7}
& \textbf{GIN} & \textbf{GraphSage} & \textbf{GCN} & \textbf{GSN} & \textbf{FragNet} & \textbf{CIN} \\
\midrule
\textbf{MPC (Ring Task)} & 7.3 & 7.3 & 7.3 & 3.6 & 3.2 & 2.9 \\
\textbf{MAE (ZINC)} & 0.53 & 0.40 & 0.37 & 0.12 & 0.078 & 0.077 \\
\bottomrule
\end{tabular}
\label{tab:ring_zinc_comparison}
\end{table}
We additionally test the ring transfer capabilities of the models on real-world graphs from the ZINC molecular regression dataset \cite{ZINC}. The ZINC dataset comprises small molecules and the benchmark task is to predict the penalized logP score which involves the number of cycles. Therefore the graphs from the ZINC dataset are well-suited to test our models cycle detection capabilities. 
For this, we first filter all graphs to contain rings of size $s$. Second, we randomly choose a node $v$ that is part of cycle $s$ for each graph. The labels are assigned in the same way as for the random regular graphs and the ER graphs.

\begin{figure}
    \centering
    \includegraphics[width=1\linewidth]{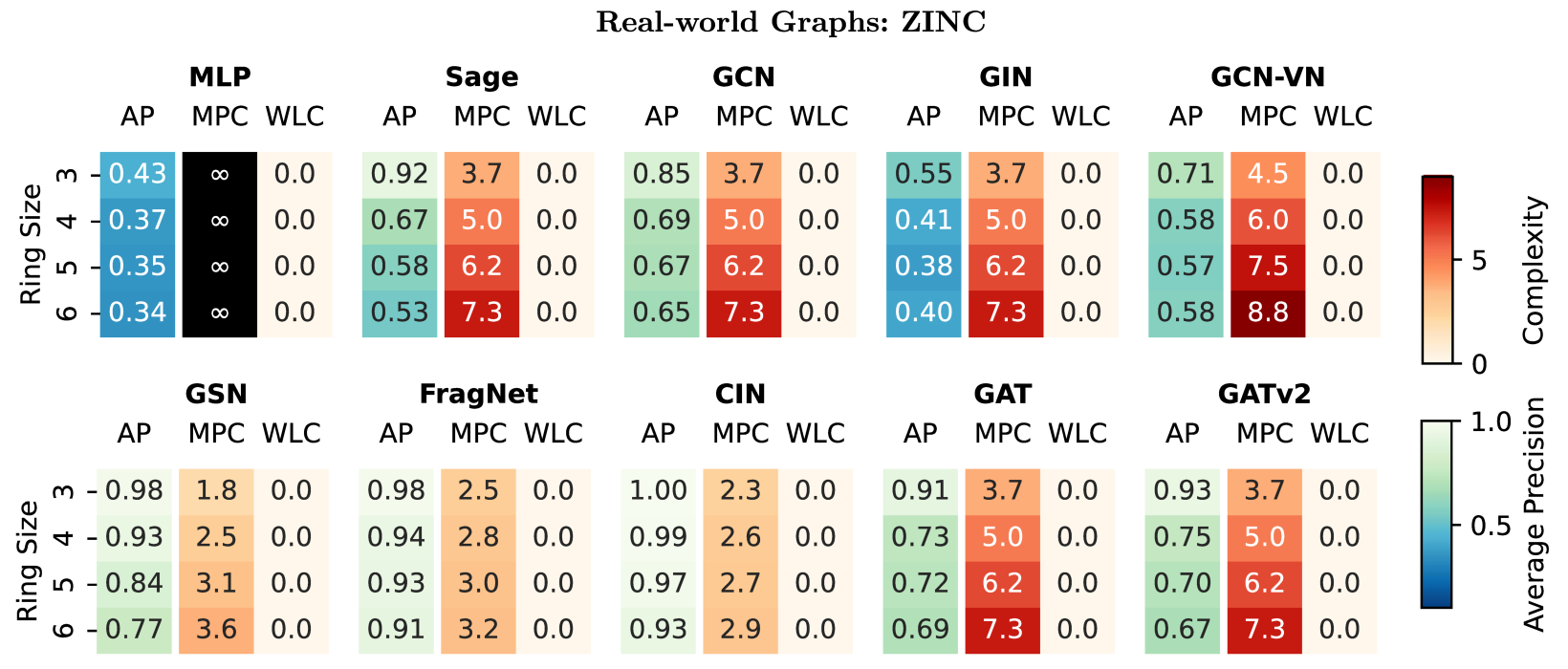}
    \caption{Test average precision compared with complexity measures \ours{} and \WLComp{} for the ring transfer task for real-world graphs from the ZINC dataset, where cycle detection is a crucial subtask (dataset size 10000, complexity values for standard MPNNs are lower bounds). \ours{} can account for performance differences in this real-world dataset that classical expressivity theory misses: The superior performance of GSN, CIN, and FragNet cannot be explained by their increased iso expressivity(\WLComp{} 0 for all MPNNs) but by their reduced \ours{} due to their cycle-oriented inductive bias.}
    \label{fig:ZINC_heatmap}
\end{figure}

\begin{figure}
    \centering
    \includegraphics[width=1\linewidth]{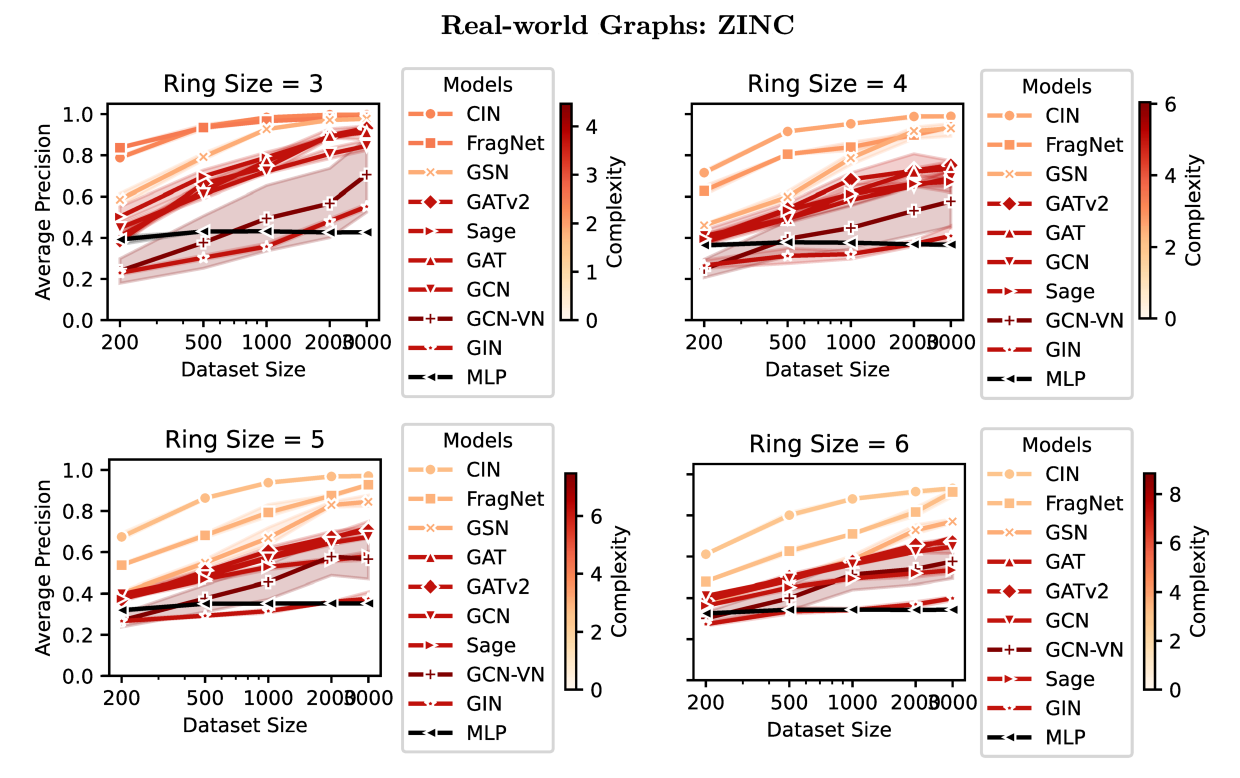}
    \caption{Test average precision compared with complexity measures \ours{} and \WLComp{} for the ring transfer task for real-world graphs from the ZINC dataset in relation to dataset size. \ours{} captures the performance advantage of the cycle-oriented GSN, CIN, and FragNet across different dataset sizes.}
    \label{fig:ZINC_line}
\end{figure}

\cref{fig:ZINC_heatmap,fig:ZINC_line} show that trends in \ours{} complexity align again with empirical performance. Again, the superior performance of GSN, CIN, and FragNet cannot be explained by their increased iso expressivity(\WLComp{} 0 for all MPNNs) but by their reduced \ours{} due to their cycle-oriented inductive bias. 

Additionally, we compare results on the synthetic ring transfer task to empirical performance on the standard ZINC molecular property prediction task (penalized logP), where ring identification is crucial for accurate predictions. As shown in \cref{tab:ring_zinc_comparison}, trends in \ours{} on the synthetic ring task match trends in empirical performance on ZINC: architectures with lower ring detection complexity (GSN, FragNet, CIN) significantly outperform standard MPNNs on both tasks. This demonstrates that MPC analysis of targeted proxy tasks can provide valuable insights into real-world performance, even when exact target functions are unknown or complexity values are computationally infeasible to derive.

\clearpage

\subsection{Simulation of Complexities}
We show in \cref{alg1} for the exemplary task of propagating information from a source node $u$ to a target node $v$ how the complexities can be efficiently simulated using Monte Carlo simulation.

\newcommand{\INPUT}{\STATE \textbf{Input:}}
\newcommand{\OUTPUT}{\STATE \textbf{Output:}}
\begin{algorithm}
\caption{Propagating Information Simulation}
\label{alg1}
\begin{algorithmic}
\State \textbf{Input:} Source node $u$, target node $v$, number of trials $T$, graph $\graphT$, edge weights $\normInf$, and maximum steps $L$
\State ~
\State $\text{success} \leftarrow 0$
\For{$t = 1$ to $T$} 
\State $\text{active} \leftarrow {u}$ 
\For{$s = 1$ to $L$}
\State $\text{newActive} \leftarrow \emptyset$
\For{$v \in \text{active}$}
\For{$u \in N_\graphT(v) \cup \{v\}$}
\If{$\text{Random}(0,1) < \normInf_{vu}$}
\State $\text{newActive} \leftarrow \text{newActive} \cup \{u\}$
\EndIf
\EndFor
\EndFor
\State $\text{active} \leftarrow \text{newActive}$
\EndFor
\If{$v \in \text{active}$}
\State $\text{success} \leftarrow \text{success} + 1$
\EndIf
\EndFor
\State ~
\State \textbf{Output:} $-\log(\text{success}/T)$
\end{algorithmic}
\end{algorithm}

\end{document}